\documentclass[table]{article}

\usepackage[final]{neurips_2021}

\usepackage{amsmath,amsfonts,bm}

\def\eqref#1{equation~\ref{#1}}

\def\1{\bm{1}}

\DeclareMathAlphabet{\mathsfit}{\encodingdefault}{\sfdefault}{m}{sl}
\SetMathAlphabet{\mathsfit}{bold}{\encodingdefault}{\sfdefault}{bx}{n}

\usepackage[utf8]{inputenc} 
\usepackage[T1]{fontenc}    
\usepackage{pgf}
\usepackage{collcell}
\usepackage{tabularx}
\usepackage{hyperref}
\usepackage{highlight}
\usepackage{url}        
\usepackage{booktabs}       
\usepackage{amsfonts}       
\usepackage{nicefrac}       
\usepackage{microtype}      
\usepackage{caption} 
\usepackage{subfig}
\usepackage{float}
\usepackage{textcomp}
\usepackage{xcolor}
\usepackage{inconsolata}
\usepackage{wrapfig}

\usepackage{xstring}

\usepackage{graphicx}
\usepackage{graphics}
\usepackage{tikz}
\usepackage{pgfplots}

\usepgfplotslibrary{groupplots}
\usetikzlibrary{pgfplots.groupplots,arrows.meta,shadows,positioning,angles,quotes}
\usetikzlibrary{matrix,chains,positioning,decorations.pathreplacing,arrows}
\usetikzlibrary{shapes.geometric}
\usetikzlibrary{positioning}

\usepgfplotslibrary{fillbetween}

\newcommand{\colorpos}{green}
\newcommand{\colorneg}{red}

\title{On The Transferability of Deep-Q Networks}

\author{
  Matthia Sabatelli\\
  Montefiore Institute\\
  Universit\'e de Li\`ege, Belgium \\
  \texttt{m.sabatelli@uliege.be} \\
  \And 
  Pierre Geurts \\
  Montefiore Institute\\
  Universit\'e de Li\`ege, Belgium \\
  \texttt{p.geurts@uliege.be} \\
}

\begin{document}

\maketitle

\begin{abstract}
	Transfer Learning (TL) is an efficient machine learning paradigm that allows overcoming some of the hurdles that characterize the successful training of deep neural networks, ranging from long training times to the needs of large datasets. While exploiting TL is a well established and successful training practice in Supervised Learning (SL), its applicability in Deep Reinforcement Learning (DRL) is rarer. In this paper, we study the level of transferability of three different variants of Deep-Q Networks on popular DRL benchmarks as well as on a set of novel, carefully designed control tasks. Our results show that transferring neural networks in a DRL context can be particularly challenging and is a process which in most cases results in negative transfer. In the attempt of understanding why Deep-Q Networks transfer so poorly, we gain novel insights into the training dynamics that characterizes this family of algorithms.    
\end{abstract}

Over the last years, the marriage between Reinforcement Learning (RL) algorithms and deep neural networks, commonly denoted as Deep Reinforcement Learning (DRL) \citep{franccois2018introduction} has gained tremendous attention \citep{henderson2018deep}. Neural networks have, in fact, proven to be extremely successful both in a model-free RL setting as in a model-based one. Large part of their success can be attributed to their ability of serving as feature extractors as well as function approximators, a property that allows them to successfully learn optimal value functions \citep{mnih2013playing,mnih2015human,van2016deep,zhao2016deep,wang2016dueling,sabatelli2020deep}, stochastic policies \citep{lillicrap2015continuous,schulman2015high,schulman2015trust,wang2016sample,mnih2016asynchronous,schulman2017proximal,haarnoja2018soft,fujimoto2018addressing}, and models of an environment \citep{ha2018world,kaiser2019model,hafner2019dream,hafner2019learning,hafner2020mastering} that is usually formalized as a Markov Decision Process (MDP) 
\cite{puterman1990markov}. Despite the many remarkable achievements, training a DRL agent is a process that can be very time-consuming. The task of solving an optimal decision making problem is,  in fact, a challenging problem of its own, which is sometimes made even more difficult by the DRL community itself, which requires DRL practitioners to test the performance of their algorithms on benchmarks that are computationally very expensive (for a position paper about this topic see \citep{obando2020revisiting}). One way of overcoming the need of individually training a DRL agent from scratch each time a new RL problem is encountered is based on Transfer Learning (TL). TL focuses on designing training strategies that allow machine learning models to retain and reuse previously learned knowledge when getting trained on new, unseen problems \citep{pan2009survey, zhuang2020comprehensive}. Within deep learning, TL is largely adopted by the Supervised Learning (SL) community \citep{huh2016makes,mormont2018comparison,sabatelli2018deep,dominguez2019transfer,vandaele2021deep,ho2021evaluation}, as it allows to train neural networks on problems that are characterized by a lack of appropriate training data or sufficient computational resources; however, typical TL approaches such as off the shelf feature extraction, or fine-tuning \citep{sharif2014cnn} have rarely been thoroughly studied from a DRL perspective. Therefore, the degree of transferability of DRL algorithms is not yet known. In this work, we focus on value-based, model-free algorithms, a family of techniques which focuses on training neural networks with the intent of learning an approximation of an optimal value function. While several of such algorithms, commonly denoted as Deep-Q Networks, exist, research studying their TL properties is, on the contrary scarce, and a clear answer to the question \textit{"How transferable are Deep-Q Networks?"} has yet to be given. In the attempt to clearly answering this question, we present the following three contributions:
\begin{itemize}
	\item We present a first \textbf{large scale empirical study} that analyses the TL properties of popular model-free DRL algorithms on the Atari Arcade Learning Environment (ALE), where we show that transferring pre-trained networks in a DRL context can be a very challenging task.
	\item We design a set of \textbf{novel, control experiments} which allows us to thoroughly characterize the TL dynamics of Deep-Q Networks.
	\item While studying Deep-Q Networks from a TL perspective, we discover \textbf{novel learning dynamics} that provide a better understanding of how this family of algorithms deals with RL tasks.  
\end{itemize}

\section{A large-scale Empirical Study}
\label{sec:empirical_study}

In this section, we carry out a large-scale TL experiment on several games from the Atari Environment (Sec. \ref{sec:atari_environments}). The experimental protocol is detailed in Sec. \ref{sec:experimental_setup_1} and results are discussed in Sec. \ref{sec:results_1}.

\subsection{The Atari Environment}
\label{sec:atari_environments}

\begin{figure}[ht]
\centering
\begin{minipage}[b]{0.35\linewidth}
\centering
\includegraphics[width=\textwidth]{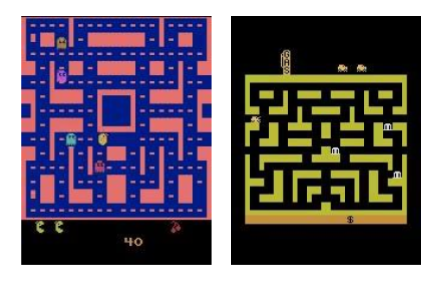}
\caption{The visually similar \texttt{Ms Pacman} and \texttt{Bank Heist} games.}
\label{fig:similar_games}
\end{minipage}
\hspace{0.5cm}
\begin{minipage}[b]{0.35\linewidth}
\centering
\includegraphics[width=\textwidth]{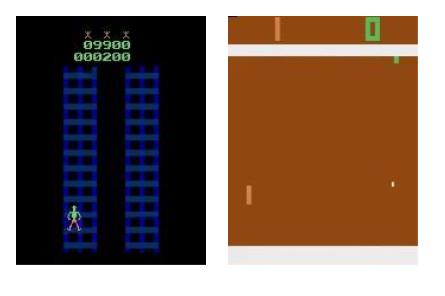}
\caption{The highly different \texttt{Crazy Climber} and \texttt{Pong} games.}
\label{fig:dissimilar_games}
\end{minipage}
\end{figure}

In this study, we use the Atari Arcade Learning Environment (ALE) \citep{bellemare2013arcade}. Next to being one of the most popular benchmarks in DRL, the ALE is particularly well suited for TL research as it allows to choose among a set of $57$ Atari games that can be used as source $\mathcal{M}_S$ and target $\mathcal{M}_T$ MDPs within a deep transfer learning setting. Since training a model-free agent on the games of the ALE is a process which can be computationally very expensive, we have carefully selected a subset of $10$ different environments. Numerous reasons guided the game selection process. First, we have selected games for which we guarantee that a model-free DRL agent can learn a good policy for. Since, as discussed by \citet{lazaric2012transfer}, one of the key requirements of TL is that of correctly identifying and transferring knowledge across source and target tasks, we naturally ensured that some knowledge coming in the form of neural network parameters representing a near-optimal value function was available for transfer. Second, while it is true that all of the selected games result in an agent that can improve its policy over time, some games were chosen because the learned policy resulted in a final performance that was not on par with that of a human expert player. This is, for example, the case of the \texttt{Frostbite} game, where the gap in performance between an agent trained with the DQV-Learning algorithm \citep{sabatelli2018deepqv} ($\approx 270$) and a human expert player ($\approx 4300$) is particularly significant. It follows that \texttt{Frostbite} is an interesting target task for transfer, as the agent's performance can potentially be improved through TL. Furthermore, we have also ensured that among the selected games, some environments are more similar to each other than others. This is, for example, the case for the \texttt{Ms. Pacman} and \texttt{Bank Heist} games which, as can be seen in Fig. \ref{fig:similar_games}, are two games where the state space is represented as a maze, and where the end goal of an agent is that of learning how to navigate it. In like manner, we have also included games that are very different from each other as is, e.g., the case for the \texttt{Crazy Climber} and \texttt{Pong} games, where it is clear from Fig. \ref{fig:dissimilar_games} that no visual similarities are shared among the two environments. Including visually similar and dissimilar games allows us to investigate whether, as is the case for supervised learning, a source task is particularly well suited for transfer if it is similar to its respective target task \citep{mensink2021factors}. 

\subsection{Experimental Setup}
\label{sec:experimental_setup_1}

We investigate the TL performance of agents that get trained with the DQV-Learning algorithm \citep{sabatelli2018deep}, and with the DDQN algorithm \citep{van2016deep}. We take models which come as pre-trained on 10 Atari games and transfer them to all remaining environments. We mostly consider the same games for both algorithms (\texttt{Bank Heist, Boxing, Crazy Climber, Fishing Derby, Frostbite, James Bond, Ms. Pacman, Pong} and \texttt{Zaxxon}) with the only exception that for DDQN \texttt{Frostbite} is replaced by \texttt{Gopher}, and \texttt{Zaxxon} is replaced by \texttt{Ice Hockey} as the \texttt{Frostbite} and \texttt{Zaxxon} DDQN agents failed to improve their policy whilst training. It is worth noting that TL is particularly easy to perform as both algorithms learn an approximation of the optimal state-action value function $Q(s,a;\theta)$ by training a convolutional neural network (see Appendix \ref{sec:experimental_setup_control} for its exact architecture and all hyper-parameter settings) directly on the images representing the state of the game. Since the state space across Atari games is always represented as an $84\times84\times4$ tensor, it is straightforward to transfer the same neural architecture among various Atari environments without needing special modifications. However, the only modification that we apply to a pre-trained network concerns its last layer responsible for estimating the different $Q$ values, which we always replace and randomly re-initialize. Following the typical deep transfer learning literature, we investigate whether in DRL it is as beneficial as it is in supervised learning to transfer a network that comes as pre-trained on $\mathcal{M}_S$ and fine-tune it on $\mathcal{M}_T$. We do this by quantitatively assessing the transfer learning benefits on each $\mathcal{M}_S/\mathcal{M}_T$ pair by computing the area ratio metric $r$ \citep{taylor2009transfer}. Specifically, given a learning curve representing the performance of an agent pre-trained on $\mathcal{M}_S$, and that of an agent that is instead trained from scratch, we compute $r$ as follows:
\begin{equation}
	r = \frac{\text{area of $\mathcal{M}_S$ $-$ area of $\mathcal{M}_T$}}{\text{area of $\mathcal{M}_T$}}.
\label{eq:area_ratio_metric}
\end{equation}

\subsection{Results}
\label{sec:results_1}

The results on each $\mathcal{M}_S/\mathcal{M}_T$ pair for both the DQV and DDQN algorithms are presented in Table \ref{tab:dqv_res}. In each cell of the tables, we report the area ratio metric defined in Eq. \ref{eq:area_ratio_metric}: the lower (resp. higher) this score, the less (resp. more) beneficial it is to transfer and fine-tune a pre-trained agent. When it comes to the DQV algorithm, we can see that, out of nine target environments, there is only one Atari game for which it is always beneficial to transfer and fine-tune a pre-trained model: \texttt{Fishing Derby}. In fact, a positive area ratio score is obtained no matter which source environment is used for pre-training, although the best results have been obtained when starting from an \texttt{Enduro} or \texttt{Pong} pre-trained network, which both resulted in an area ratio score of $\approx 0.72$. Positive transfer can also be observed on the \texttt{Frostbite} and \texttt{James Bond} games, but only for a limited number of source games. For example, a \texttt{Bank Heist} pre-trained agent transfers well to both target games as it obtains an area ratio score of $0.729$ and $0.973$ respectively, but the same cannot be said for an \texttt{Enduro} pre-trained network, which on \texttt{Frostbite} results in absent transfer (the area ratio score is, in fact, $-0.017$), and yields negative transfer on \texttt{James Bond} ($r=-0.41$). We can also observe that there are environments where it is surprisingly never beneficial to transfer and fine-tune a pre-trained agent. This is, for example, the case for the \texttt{Bank Heist} and \texttt{Pong} games, where independently from which source game $\mathcal{M}_S$ is used for pre-training, a negative area ratio score is always obtained. Furthermore, it can also be observed that transfer learning across environments is not symmetric, as one source game $\mathcal{M}_S$ can result in positive transfer when it gets transferred to a certain target game $\mathcal{M}_T$, but the same outcome is not obtained when transfer is performed in the opposite direction. As an example we can consider the \texttt{Boxing/Fishing Derby} games: positive transfer is obtained when transferring from \texttt{Boxing}$\rightarrow$\texttt{Fishing Derby} ($r=0.552$), but negative transfer is obtained when transferring from \texttt{Fishing Derby} $\rightarrow$ \texttt{Boxing} ($r=-0.893$). When it comes to the DDQN algorithm, similar conclusions can be drawn: we can again observe that there are only very few cases for which it is beneficial to transfer and fine-tune a pre-trained DRL agent. Examples of such cases are networks that are pre-trained on \texttt{Ice Hockey} and \texttt{James Bond} which get transferred to \texttt{Boxing} ($r=0.245$ and $r=0.232$ respectively), or \texttt{Boxing} and \texttt{Enduro} models that get transferred to \texttt{Pong} ($r=0.936$ and $r=0.248$). \texttt{Bank Heist} and \texttt{Pong} are again the two target environments for which most of the transferred source models resulted in negative transfer, while differently from the experiments performed with the DQV algorithm, this time no positive transfer can be observed on \texttt{Fishing Derby}. Overall, the process of fine-tuning a pre-trained DDQN agent mostly results in absent transfer, as can be observed by the area ratio scores obtained on \texttt{Enduro, Fishing Derby, Gopher} and \texttt{Ice Hockey} which are all $\approx 0$ on average.

\begin{table}[t!]
	\caption{The results obtained when fine-tuning ten different pre-trained agents (rows) on nine other Atari games (columns), with DQV (top table) and DDQN (bottom table). Positive values (in \colorpos) represent positive transfer, while negative values (in \colorneg) represent negative transfer. The darker the color, the higher the absolute value of the area ratio score.} 
	{

\newcommand*{\MinNumber}{-1.0}%
\newcommand*{\MidNumber}{0.0}%
\newcommand*{\MaxNumber}{1.0}%

\newcommand{\ApplyGradient}[1]{%
  \IfDecimal{#1}{
    \ifdim #1 pt > \MidNumber pt%
    \pgfmathsetmacro{\PercentColor}{max(min(100.0*(#1-\MidNumber)/(\MaxNumber-\MidNumber),100.0),0.0)}%
    \edef\x{\noexpand\cellcolor{\colorpos!\PercentColor!white}}\x\textcolor{black}{#1}%
    \else%
    \pgfmathsetmacro{\PercentColor}{max(min(100.0*(\MidNumber-#1)/(\MidNumber-\MinNumber),100.0),0.0)}%
    \edef\x{\noexpand\cellcolor{\colorneg!\PercentColor!white}}\x\textcolor{black}{#1}%
    \fi%
    }{#1}
}

\newcolumntype{R}{>{\collectcell\ApplyGradient}{c}<{\endcollectcell}}

\resizebox{\columnwidth}{!}{%
\begin{tabular}{c|R|R|R|R|R|R|R|R|R|R}
\hline 
{DQV} & BankHeist & Boxing & CrazyClimber & Enduro & FishingDerby & Frostbite & JamesBond & MsPacman &   Pong & Zaxxon \\
\hline \hline 
BankHeist    &     -      & -0.019 &           -1 & -0.317 &          0.5 &     0.729 &     0.973 &   -0.089 & -1.238 & -0.998 \\
Boxing       &    -0.494 &    -    &       -0.278 & -0.852 &        0.552 &     -0.01 &     0.247 &   -0.184 & -0.841 & -0.999 \\
CrazyClimber &    -0.569 & -0.261 &           -   & -0.593 &         0.19 &     0.277 &     0.621 &   -0.111 & -1.206 & -0.178 \\
Enduro       &    -0.571 & -0.018 &        -0.25 &     -   &        0.726 &    -0.017 &     -0.41 &    -0.08 & -0.466 & -0.164 \\
FishingDerby &        -1 & -0.893 &       -0.093 &  -0.45 &            -  &     0.068 &     0.197 &   -0.136 & -3.083 & -0.999 \\
Frostbite    &    -0.933 &  0.024 &           -1 & -0.348 &        0.222 &        -   &     0.569 &    0.009 & -0.663 & -0.076 \\
JamesBond    &    -0.123 & -0.106 &       -0.131 & -0.033 &        0.519 &     0.262 &         -  &    0.218 & -1.329 &     -1 \\
MsPacman     &    -0.985 & -0.219 &       -0.012 & -0.494 &          0.6 &     0.346 &     0.398 &        -  & -1.646 & -0.997 \\
Pong         &        -1 & -0.083 &       -0.428 & -0.476 &        0.725 &    -0.024 &     0.896 &    0.123 &    -    & -0.729 \\
Zaxxon       &     -0.76 & -0.028 &        0.037 & -0.116 &        0.385 &      0.16 &    -0.253 &     0.06 & -1.602 &   -     \\
\end{tabular}%
}

}
	\label{tab:dqv_res}

        ~\\
	{\newcommand*{\MinNumber}{-1.0}%
\newcommand*{\MidNumber}{0.0}%
\newcommand*{\MaxNumber}{1.0}%

\newcommand{\ApplyGradient}[1]{%
  \IfDecimal{#1}{
    \ifdim #1 pt > \MidNumber pt%
    \pgfmathsetmacro{\PercentColor}{max(min(100.0*(#1-\MidNumber)/(\MaxNumber-\MidNumber),100.0),0.0)}%
    \edef\x{\noexpand\cellcolor{\colorpos!\PercentColor!white}}\x\textcolor{black}{#1}%
    \else%
    \pgfmathsetmacro{\PercentColor}{max(min(100.0*(\MidNumber-#1)/(\MidNumber-\MinNumber),100.0),0.0)}%
    \edef\x{\noexpand\cellcolor{\colorneg!\PercentColor!white}}\x\textcolor{black}{#1}%
    \fi%
    }{#1}
}

\newcolumntype{R}{>{\collectcell\ApplyGradient}{c}<{\endcollectcell}}

\resizebox{\columnwidth}{!}{%
\begin{tabular}{c|R|R|R|R|R|R|R|R|R|R}
\hline 
{DDQN} & BankHeist & Boxing & CrazyClimber & Enduro & FishingDerby & Gopher & IceHockey & Jamesbond & MsPacman & Pong \\
\hline\hline
BankHeist    &       -                    &                  0.121 &                       -0.378 &                 -0.006 &                       -0.107 &                  0.042 &                    -0.006 &                    -0.058 &                    0.001 &               -3.013 \\
Boxing       &                    -0.316 &     -                   &                       -0.104 &                     -0 &                        0.038 &                   0.06 &                     0.015 &                    -0.225 &                   -0.027 &                0.936 \\
CrazyClimber &                    -0.192 &                 -0.487 &               -               &                 -0.012 &                       -0.084 &                  0.016 &                     0.015 &                     0.016 &                   -0.015 &                -2.64 \\
Enduro       &                    -0.296 &                  0.193 &                       -0.167 &       -                 &                        0.039 &                   0.03 &                     0.019 &                    -0.235 &                   -0.039 &                0.248 \\
FishingDerby &                    -0.212 &                 -0.545 &                           -1 &                 -0.085 &         -                     &                  0.016 &                     0.001 &                    -0.055 &                   -0.026 &               -0.935 \\
Gopher       &                    -0.466 &                  0.044 &                       -0.108 &                 -0.005 &                        0.007 &            -            &                    -0.005 &                    -0.094 &                    -0.02 &               -1.816 \\
IceHockey    &                    -0.046 &                  0.245 &                       -0.067 &                  0.014 &                       -0.178 &                  0.072 &            -               &                     0.037 &                   -0.015 &                0.112 \\
Jamesbond    &                    -0.145 &                  0.232 &                       -0.064 &                  0.005 &                       -0.267 &                  0.031 &                    -0.092 &      -                     &                   -0.006 &               -1.578 \\
MsPacman     &                    -0.173 &                 -1.179 &                       -0.129 &                  -0.06 &                        0.003 &                 -0.019 &                     0.007 &                     0.071 &       -                   &               -2.774 \\
Pong         &                    -0.127 &                  0.028 &                        -0.12 &                   0.01 &                        0.037 &                  0.042 &                     0.002 &                    -0.174 &                   -0.006 &             -         \\
\end{tabular}%
}
}
	\label{tab:ddqn_res}
\end{table}

\section{Control Experiments}
\label{sec:control_experiments}
The results presented in the previous section seem to be questioning the level of transferability of DRL agents. In fact, the training strategy of fine-tuning a pre-trained model on $\mathcal{M}_T$ does not result in the same type of performance gains that have been extensively observed in a supervised learning context. To better characterize their TL properties, we have designed a set of simple control experiments that allow us to examine their transfer learning behavior in training conditions that do not require extraordinarily long training times for learning an optimal policy.   

\subsection{The Catch Environments}
\label{sec:catch_environments}

To this end, we have implemented four different versions of the \texttt{Catch} game, a simple RL task that was first presented by \citet{mnih2014recurrent}, and that has been widely used within the literature for investigating the performance of DRL algorithms in a fast, and computationally less expensive manner than the one required by the \texttt{Atari} games \citep{vanjos2018deep, aittahar2020empirical}. In the game of \texttt{Catch}, an agent controls a paddle at the bottom of the environment, represented by a $21 \times 21$ grid, and has to catch a ball falling from top to bottom, which can potentially bounce off walls.
At each time step, the agent can choose between three actions: move the paddle one pixel to the right, move it to the left, or keep it in the same position in the grid. An RL episode ends either when the agent manages to catch the ball, in which case it receives a reward of $1$, or when it misses the ball, which naturally results in a reward of $0$. Following the design choices presented in \citep{vanjos2017deep}, we model the ball to have vertical speed of $v_y=-1 \: cell/s$ and horizontal speed of $v_x \in \{-2,-1,0,1,2\}$. From now on, we will refer to this version of the game as \texttt{Catch-v0}, as it is the most basic and simplest form of the game that will be used throughout our experiments. Next to \texttt{Catch-v0} we have implemented three slightly different and arguably more complex versions of the game as well: \texttt{Catch-v1}, where we increased the complexity of the game by reducing the size of the paddle that the agent controls. While for \texttt{Catch-v0} its size is of five pixels, in \texttt{Catch-v1} it is of two pixels, therefore requiring the agent to be more precise if it wants to successfully catch the falling ball. The second alternative version of \texttt{Catch} is \texttt{Catch-v2}. In this case, the dynamics of the game are identical to the ones that define \texttt{Catch-v0}; however, the way the $21 \times 21$ grid is represented changes. While in \texttt{Catch-v0} as well as in \texttt{Catch-v1} the state is represented by a binary grid where all pixels, but the ones representing the paddle and the ball have a value of $0$, in \texttt{Catch-v2} the cells around the paddle and the ball can have a random value between $0$ and $255$. 
\begin{figure}[ht]
\begin{minipage}{0.5\textwidth}
	\centering
\includegraphics[width=5.5cm]{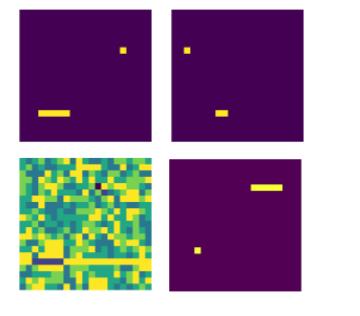}
\end{minipage}%
\begin{minipage}{0.5\textwidth}
	\centering
	\begin{tikzpicture}[scale=0.65]
\begin{axis}
[	grid style={dashed,gray},
	grid = both, 
	tick style=black,
  	xlabel=Epochs,
	ylabel = $\%$ of caught balls,
	legend pos=outer north east,
]

\addlegendentry{\texttt{Catch-v0}}

\addplot[blue, ultra thick] table[x=epochs,y=catchv0] {./Results/logs/catch_baselines.dat};
\addplot[name path=upper,draw=none, forget plot] table[x=epochs,y expr=\thisrow{catchv0}+\thisrow{stdv0}] {./Results/logs/catch_baselines.dat};
\addplot[name path=lower,draw=none, forget plot] table[x=epochs,y expr=\thisrow{catchv0}-\thisrow{stdv0}] {./Results/logs/catch_baselines.dat};
\addplot[fill=blue!10, forget plot] fill between[of=upper and lower];

\addlegendentry{\texttt{Catch-v1}}
\addplot[ultra thick, yellow] table[x=epochs,y=catchv3] {./Results/logs/catch_baselines.dat};
\addplot [name path=upper,draw=none, forget plot] table[x=epochs,y expr=\thisrow{catchv3}+\thisrow{stdv3}] {./Results/logs/catch_baselines.dat};
\addplot [name path=lower,draw=none, forget plot] table[x=epochs,y expr=\thisrow{catchv3}-\thisrow{stdv3}] {./Results/logs/catch_baselines.dat};
\addplot [fill=yellow!10, forget plot] fill between[of=upper and lower];

\addlegendentry{\texttt{Catch-v2}}
\addplot[ultra thick, red] table[x=epochs,y=catchv1] {./Results/logs/catch_baselines.dat};
\addplot[name path=upper,draw=none, forget plot] table[x=epochs,y expr=\thisrow{catchv1}+\thisrow{stdv1}] {./Results/logs/catch_baselines.dat};
\addplot[name path=lower,draw=none, forget plot] table[x=epochs,y expr=\thisrow{catchv1}-\thisrow{stdv1}] {./Results/logs/catch_baselines.dat};
\addplot[fill=red!10, forget plot] fill between[of=upper and lower];

\addlegendentry{\texttt{Catch-v3}}
\addplot[ultra thick, green] table[x=epochs,y=catchv2] {./Results/logs/catch_baselines.dat};
\addplot [name path=upper,draw=none, forget plot] table[x=epochs,y expr=\thisrow{catchv2}+\thisrow{stdv2}] {./Results/logs/catch_baselines.dat};
\addplot [name path=lower,draw=none, forget plot] table[x=epochs,y expr=\thisrow{catchv2}-\thisrow{stdv2}] {./Results/logs/catch_baselines.dat};
\addplot [fill=green!10, forget plot] fill between[of=upper and lower];

\end{axis}
\end{tikzpicture}
\end{minipage}
\caption{Image on the left: the four different versions of the \texttt{Catch} environment. In clockwise order: \texttt{Catch-v0, Catch-v1, Catch-v3} and \texttt{Catch-v2}. Image on the right: learning curves obtained by a DQN agent that is trained from scratch on the aforementioned \texttt{Catch} versions. Shaded areas correspond to $\pm1$ std. obtained over 5 different random seeds.}
\label{fig:catch_baselines}
\end{figure}
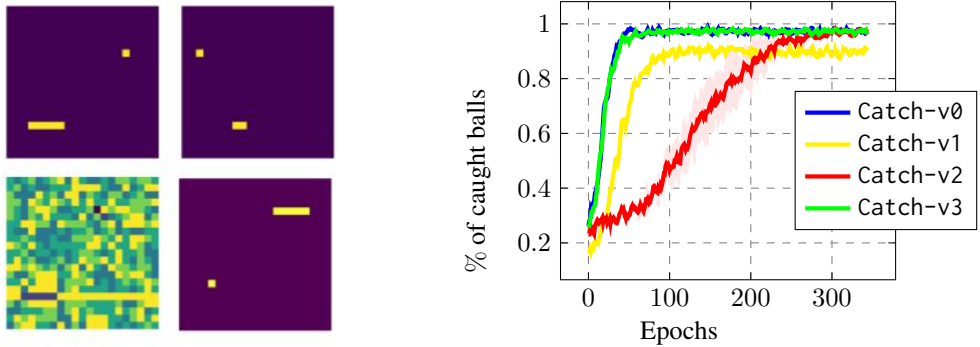
This design choice makes it much harder for a convolutional network to correctly locate and identify the position of the paddle and of the falling ball and makes \texttt{Catch-v2} the arguably most complex version among the different \texttt{Catch} environments. Lastly, we have implemented \texttt{Catch-v3}, a version of \texttt{Catch} which is identical to the one that is modeled by \texttt{Catch-v0} with the main difference that the representation of the state is now mirrored, therefore requiring the agent to look at different parts of the grid if it wants to locate the paddle, and understand that the ball is unnaturally moving from the bottom to the top. For an impression of all four \texttt{Catch} versions see the left image of Fig. \ref{fig:catch_baselines}.
Given the overall simplicity of the different \texttt{Catch} environments, we now train a DQN agent instead of the arguably more complex DQV and DDQN agents that we considered in Sec. \ref{sec:empirical_study}. As we can see from the results reported in the right plot of Fig. \ref{fig:catch_baselines}, averaged over five different runs, the agent is able to successfully learn a near optimal policy for all \texttt{Catch} versions. When it comes to \texttt{Catch-v0}, \texttt{Catch-v2} and \texttt{Catch-v3} we can observe that by the end of training, the agent is able to catch $\approx 100\%$ of the falling balls, whereas its performance is slightly worse ($\approx 90\%$) when it comes to \texttt{Catch-v1} \footnote{Please note that the performance on \texttt{Catch-v1} can be improved by increasing the complexity of the DQN agent by adding one more convolutional layer. However, as the goal is to transfer the same type of model across \texttt{Catch} games, we did not modify the architecture of the DQN agent, at the cost of having a slightly worse performing model.}. We can also observe that among the different \texttt{Catch} versions, \texttt{Catch-v0} and \texttt{Catch-v3} appear to be the easiest ones, as the agent requires significantly less training episodes to converge when compared to \texttt{Catch-v1} and \texttt{Catch-v2}. Furthermore, in line with the explanation presented beforehand, our results also confirm the hypothesis that \texttt{Catch-v2} is the overall most complicated \texttt{Catch} version, as learning requires significantly more, and potentially unstable, training epochs.

\subsection{From one Catch to Another}
\label{sec:from_one_catch_to_another}
\begin{figure}[ht!]
\resizebox{0.45\textwidth}{!}{  \centering
\begin{tikzpicture}
\begin{axis}
[	
	name=ax1,	
	grid style={dashed,gray},
	grid = both, 
	tick style=black,
  	xlabel=Epochs,
	ylabel = $\%$ of caught balls,
	title = \texttt{Catch-v0},
	legend pos=south east,	
]

\addplot[blue, ultra thick] table[x=epochs,y=target_catch] {./Results/logs/catch_v0.dat};
\addplot [name path=upper,draw=none, forget plot] table[x=epochs,y expr=\thisrow{target_catch}+\thisrow{std_target_catch}] {./Results/logs/catch_v0.dat};
\addplot [name path=lower,draw=none, forget plot] table[x=epochs,y expr=\thisrow{target_catch}-\thisrow{std_target_catch}] {./Results/logs/catch_v0.dat};
\addplot [fill=blue!10, forget plot] fill between[of=upper and lower];

\addplot[ultra thick, red] table[x=epochs,y=Catch_v2] {./Results/logs/catch_v0.dat};
\addplot [name path=upper,draw=none, forget plot] table[x=epochs,y expr=\thisrow{Catch_v2}+\thisrow{Catch_v2_std}] {./Results/logs/catch_v0.dat};
\addplot [name path=lower,draw=none, forget plot] table[x=epochs,y expr=\thisrow{Catch_v2}-\thisrow{Catch_v2_std}] {./Results/logs/catch_v0.dat};
\addplot [fill=red!10, forget plot] fill between[of=upper and lower];

\addplot[ultra thick, green] table[x=epochs,y=Catch_v3] {./Results/logs/catch_v0.dat};
\addplot [name path=upper,draw=none, forget plot] table[x=epochs,y expr=\thisrow{Catch_v3}+\thisrow{Catch_v3_std}] {./Results/logs/catch_v0.dat};
\addplot [name path=lower,draw=none, forget plot] table[x=epochs,y expr=\thisrow{Catch_v3}-\thisrow{Catch_v3_std}] {./Results/logs/catch_v0.dat};
\addplot [fill=green!10, forget plot] fill between[of=upper and lower];

\addplot[ultra thick, yellow] table[x=epochs,y=Catch_v4] {./Results/logs/catch_v0.dat};
\addplot [name path=upper,draw=none, forget plot] table[x=epochs,y expr=\thisrow{Catch_v4}+\thisrow{Catch_v4_std}] {./Results/logs/catch_v0.dat};
\addplot [name path=lower,draw=none, forget plot] table[x=epochs,y expr=\thisrow{Catch_v4}-\thisrow{Catch_v4_std}] {./Results/logs/catch_v0.dat};
\addplot [fill=yellow!10, forget plot] fill between[of=upper and lower];

\end{axis}

\begin{axis}
[	
	at={(ax1.south east)},
	xshift=1cm,
	grid style={dashed,gray},
	grid = both, 
	tick style=black,
  	xlabel=Epochs,
	title = \texttt{Catch-v1},
	legend pos=south east,
]


\addplot[yellow, ultra thick] table[x=epochs,y=target_catch] {./Results/logs/catch_v1.dat};
\addplot [name path=upper,draw=none, forget plot] table[x=epochs,y expr=\thisrow{target_catch}+\thisrow{std_target_catch}] {./Results/logs/catch_v1.dat};
\addplot [name path=lower,draw=none, forget plot] table[x=epochs,y expr=\thisrow{target_catch}-\thisrow{std_target_catch}] {./Results/logs/catch_v1.dat};
\addplot [fill=yellow!10, forget plot] fill between[of=upper and lower];

\addplot[ultra thick, red] table[x=epochs,y=Catch_v0] {./Results/logs/catch_v1.dat};
\addplot [name path=upper,draw=none, forget plot] table[x=epochs,y expr=\thisrow{Catch_v0}+\thisrow{Catch_v0_std}] {./Results/logs/catch_v1.dat};
\addplot [name path=lower,draw=none, forget plot] table[x=epochs,y expr=\thisrow{Catch_v0}-\thisrow{Catch_v0_std}] {./Results/logs/catch_v1.dat};
\addplot [fill=red!10, forget plot] fill between[of=upper and lower];

\addplot[ultra thick, green] table[x=epochs,y=Catch_v2] {./Results/logs/catch_v1.dat};
\addplot [name path=upper,draw=none, forget plot] table[x=epochs,y expr=\thisrow{Catch_v2}+\thisrow{Catch_v2_std}] {./Results/logs/catch_v1.dat};
\addplot [name path=lower,draw=none, forget plot] table[x=epochs,y expr=\thisrow{Catch_v2}-\thisrow{Catch_v2_std}] {./Results/logs/catch_v1.dat};
\addplot [fill=green!10, forget plot] fill between[of=upper and lower];

\addplot[ultra thick, blue] table[x=epochs,y=Catch_v3] {./Results/logs/catch_v1.dat};
\addplot [name path=upper,draw=none, forget plot] table[x=epochs,y expr=\thisrow{Catch_v3}+\thisrow{Catch_v3_std}] {./Results/logs/catch_v1.dat};
\addplot [name path=lower,draw=none, forget plot] table[x=epochs,y expr=\thisrow{Catch_v3}-\thisrow{Catch_v3_std}] {./Results/logs/catch_v1.dat};
\addplot [fill=blue!10, forget plot] fill between[of=upper and lower];

\end{axis}

\begin{axis}
[	at={(ax1.south east)},
	xshift=9cm,
    grid style={dashed,gray},
	grid = both, 
	tick style=black,
  	xlabel=Epochs,
	legend pos=south east,	
	title = \texttt{Catch-v2}
]

\addplot[red, ultra thick] table[x=epochs,y=target_catch] {./Results/logs/catch_v2.dat};
\addplot [name path=upper,draw=none, forget plot] table[x=epochs,y expr=\thisrow{target_catch}+\thisrow{std_target_catch}] {./Results/logs/catch_v2.dat};
\addplot [name path=lower,draw=none, forget plot] table[x=epochs,y expr=\thisrow{target_catch}-\thisrow{std_target_catch}] {./Results/logs/catch_v2.dat};
\addplot [fill=red!10, forget plot] fill between[of=upper and lower];

\addplot[ultra thick, blue] table[x=epochs,y=Catch_v0] {./Results/logs/catch_v2.dat};
\addplot [name path=upper,draw=none, forget plot] table[x=epochs,y expr=\thisrow{Catch_v0}+\thisrow{Catch_v0_std}] {./Results/logs/catch_v2.dat};
\addplot [name path=lower,draw=none, forget plot] table[x=epochs,y expr=\thisrow{Catch_v0}-\thisrow{Catch_v0_std}] {./Results/logs/catch_v2.dat};
\addplot [fill=blue!10, forget plot] fill between[of=upper and lower];

\addplot[ultra thick, green] table[x=epochs,y=Catch_v3] {./Results/logs/catch_v2.dat};
\addplot [name path=upper,draw=none, forget plot] table[x=epochs,y expr=\thisrow{Catch_v3}+\thisrow{Catch_v3_std}] {./Results/logs/catch_v2.dat};
\addplot [name path=lower,draw=none, forget plot] table[x=epochs,y expr=\thisrow{Catch_v3}-\thisrow{Catch_v3_std}] {./Results/logs/catch_v2.dat};
\addplot [fill=green!10, forget plot] fill between[of=upper and lower];

\addplot[ultra thick, yellow] table[x=epochs,y=Catch_v4] {./Results/logs/catch_v2.dat};
\addplot [name path=upper,draw=none, forget plot] table[x=epochs,y expr=\thisrow{Catch_v4}+\thisrow{Catch_v4_std}] {./Results/logs/catch_v2.dat};
\addplot [name path=lower,draw=none, forget plot] table[x=epochs,y expr=\thisrow{Catch_v4}-\thisrow{Catch_v4_std}] {./Results/logs/catch_v2.dat};
\addplot [fill=yellow!10, forget plot] fill between[of=upper and lower];

\end{axis}

\begin{axis}
[	
    at={(ax1.south east)},
	xshift=17cm,
    grid style={dashed,gray},
	grid = both, 
	tick style=black,
  	xlabel=Epochs,
	title= \texttt{Catch-v3}
]

\addlegendentry{\texttt{Catch-v0}}
\addplot[ultra thick, blue] table[x=epochs,y=Catch_v0] {./Results/logs/catch_v3.dat};
\addplot [name path=upper,draw=none, forget plot] table[x=epochs,y expr=\thisrow{Catch_v0}+\thisrow{Catch_v0_std}] {./Results/logs/catch_v3.dat};
\addplot [name path=lower,draw=none, forget plot] table[x=epochs,y expr=\thisrow{Catch_v0}-\thisrow{Catch_v0_std}] {./Results/logs/catch_v3.dat};
\addplot [fill=blue!10, forget plot] fill between[of=upper and lower];

\addlegendentry{\texttt{Catch-v1}}
\addplot[ultra thick, yellow] table[x=epochs,y=Catch_v4] {./Results/logs/catch_v3.dat};
\addplot [name path=upper,draw=none, forget plot] table[x=epochs,y expr=\thisrow{Catch_v4}+\thisrow{Catch_v4_std}] {./Results/logs/catch_v3.dat};
\addplot [name path=lower,draw=none, forget plot] table[x=epochs,y expr=\thisrow{Catch_v4}-\thisrow{Catch_v4_std}] {./Results/logs/catch_v3.dat};
\addplot [fill=yellow!10, forget plot] fill between[of=upper and lower];

\addlegendentry{\texttt{Catch-v2}}
\addplot[ultra thick, red] table[x=epochs,y=Catch_v2] {./Results/logs/catch_v3.dat};
\addplot [name path=upper,draw=none, forget plot] table[x=epochs,y expr=\thisrow{Catch_v2}+\thisrow{Catch_v2_std}] {./Results/logs/catch_v3.dat};
\addplot [name path=lower,draw=none, forget plot] table[x=epochs,y expr=\thisrow{Catch_v2}-\thisrow{Catch_v2_std}] {./Results/logs/catch_v3.dat};
\addplot [fill=red!10, forget plot] fill between[of=upper and lower];

\addlegendentry{\texttt{Catch-v3}}
\addplot[green, ultra thick] table[x=epochs,y=target_catch] {./Results/logs/catch_v3.dat};
\addplot [name path=upper,draw=none, forget plot] table[x=epochs,y expr=\thisrow{target_catch}+\thisrow{std_target_catch}] {./Results/logs/catch_v3.dat};
\addplot [name path=lower,draw=none, forget plot] table[x=epochs,y expr=\thisrow{target_catch}-\thisrow{std_target_catch}] {./Results/logs/catch_v3.dat};
\addplot [fill=green!10, forget plot] fill between[of=upper and lower];

\end{axis}

\end{tikzpicture}%
}
\caption{The results obtained after using a pre-trained \texttt{Catch} agent and fine-tuning it on a different \texttt{Catch} version. We can observe that despite all \texttt{Catch} versions being very similar no positive transfer is ever observed, as a model trained from scratch always outperforms a pre-trained, fine-tuned network.}
\label{fig:catch_transfer}
\end{figure}
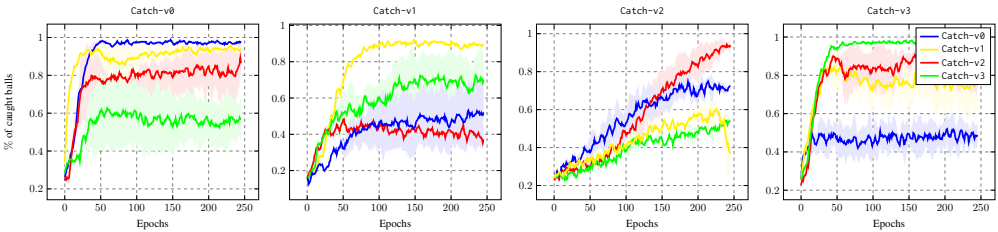

We now replicate the TL study presented in Sec. \ref{sec:empirical_study} on the aforementioned \texttt{Catch} environments, with the hope of identifying why the process of fine-tuning a pre-trained convolutional neural network in a DRL context, seems to not be as beneficial as it is in the supervised learning one. Our goal is to find at least one pair of \texttt{Catch} environments which results in positive transfer, and to then potentially identify some properties within the different \texttt{Catch} versions that could also hold for the pairs of \texttt{Atari} games which have yielded positive transfer in Table \ref{tab:dqv_res}. We formulate two hypothesis, based on which, we expect to experimentally observe positive transfer. First, we foresee that positive transfer will happen for all possible \texttt{Catch} combinations, as in the end the source MDP $\mathcal{M}_S$ and the target MDP $\mathcal{M}_T$ do not significantly differ from each other: in fact, the main task across different \texttt{Catch} versions remains that of catching a falling ball; the action space is identical; and so is the reward function that always returns a value of $1$ when the agent succeeds in catching the falling ball. This hypothesis is also motivated by supervised learning research which shows that the higher the similarity between the source domain $\mathcal{D}_S$ and the target domain $\mathcal{D}_T$, the better the performance of a transferred pre-trained network \citep{sabatelli2018deep}.
Second, in the case the previous hypothesis will not be empirically supported, we expect to at least observe positive transfer when using a model that comes as pre-trained either on \texttt{Catch-v1} or on \texttt{Catch-v2}. In fact, as described above and also shown by the performance reported in Fig. \ref{fig:catch_baselines}, these are the two most complicated versions of the \texttt{Catch} environment. In supervised learning, one of the main factors that makes a certain source task $\mathcal{T}_S$ good for transferring is its complexity \citep{mensink2021factors}, which is usually defined in terms of dataset size and number of classes to classify, therefore we expect the complexity of the source game to play an important role within DRL as well. Similarly to what we did for DQV and DDQN in Sec. \ref{sec:empirical_study}, we take the four different models that have been trained from scratch on their respective \texttt{Catch} version, randomly re-initialize their last layer (responsible for estimating the different state-action values $Q(s,a)$) and fully fine-tune the pre-trained network on the three remaining \texttt{Catch} environments.
\begin{wraptable}{r}{8.5cm}
	\centering
	\caption{The area ratio obtained after fine-tuning a pre-trained DQN agent on the different \texttt{Catch} environments. We can see that no matter which source game is used for pre-training, transfer learning surprisingly never results in positive transfer.}
	\newcommand*{\MinNumber}{-0.49}%
\newcommand*{\MidNumber}{0.0}%
\newcommand*{\MaxNumber}{0.49}%

\newcommand{\ApplyGradient}[1]{%
  \IfDecimal{#1}{
    \ifdim #1 pt > \MidNumber pt%
    \pgfmathsetmacro{\PercentColor}{max(min(100.0*(#1-\MidNumber)/(\MaxNumber-\MidNumber),100.0),0.0)}%
    \edef\x{\noexpand\cellcolor{\colorpos!\PercentColor!white}}\x\textcolor{black}{#1}%
    \else%
    \pgfmathsetmacro{\PercentColor}{max(min(100.0*(\MidNumber-#1)/(\MidNumber-\MinNumber),100.0),0.0)}%
    \edef\x{\noexpand\cellcolor{\colorneg!\PercentColor!white}}\x\textcolor{black}{#1}%
    \fi%
    }{#1}
}

\newcolumntype{R}{>{\collectcell\ApplyGradient}{c}<{\endcollectcell}}
\begin{tabular}{c|R|R|R|R}
\hline 
{} & Catch-v0 & Catch-v1 & Catch-v2 & Catch-v3 \\ \hline\hline
Catch-v0 &       -   &   -0.026 &   -0.486 &   -0.479 \\
Catch-v1 &    -0.16 &       -   &   -0.121 &   -0.248 \\
Catch-v2 &   -0.406 &   -0.313 &      -    &   -0.465 \\
Catch-v3 &   -0.016 &    -0.24 &   -0.179 &       -   \\
\end{tabular}

	\label{tab:catch_tl_area_ratio}
\end{wraptable}
The results of this study are reported in Fig. \ref{fig:catch_transfer}, where, from left to right, we show the performance that is obtained when considering \texttt{Catch-v0}, \texttt{Catch-v1}, \texttt{Catch-v2} and \texttt{Catch-v3} as target MDP $\mathcal{M}_T$. The performance of each transferred network is compared against the performance that is obtained after training a DQN agent from scratch which matches with the results reported in Fig. \ref{fig:catch_baselines}.
Surprisingly we found that fine-tuning a pre-trained DQN agent never resulted in positive transfer learning. This can clearly be seen in all plots represented in Fig. \ref{fig:catch_transfer} and by the results reporting the area ratio metric in Table \ref{tab:catch_tl_area_ratio}. The only case where starting from a pre-trained network appeared to be at least in part beneficial is represented by the first plot of Fig. \ref{fig:catch_transfer} when \texttt{Catch-v1} is considered as source MDP $\mathcal{M}_S$. In this case we can in fact observe some learning speed improvements within the first $25$ learning epochs. This is not surprising as an agent which is able to catch a ball with a small paddle (as defined by the game \texttt{Catch-v1}) should in principle also be able to do this when the size of its paddle is larger (which is the case for \texttt{Catch-v0}). What is more surprising, however, is that while training progresses we see that the performance of a \texttt{Catch-v1} pre-trained model starts deteriorating and that this model barely converges to the same performance that is obtained by a model trained from scratch. When it comes to all other $\mathcal{M}_S/\mathcal{M}_T$ pairs we see that pre-trained networks always perform significantly worse than randomly initialized models trained from scratch, with some extreme cases, as the one reported in the last plot of Fig. \ref{fig:catch_transfer}, where a \texttt{Catch-v0} pre-trained agent is barely able to improve its policy over time at all. These results invalidate our two hypotheses mentioned above as they clearly show that positive transfer in DRL does not arise when $\mathcal{M}_S$ and $\mathcal{M}_T$ are similar, nor when $\mathcal{M}_S$ is more complex than $\mathcal{M}_T$. 

\subsection{Self-Transfer}
\label{sec:self_transfer}

\begin{figure}[ht!]
\resizebox{0.45\textwidth}{!}{
\begin{tikzpicture}
\begin{axis}
[	
	name=ax1,	
	grid style={dashed,gray},
	grid = both, 
	tick style=black,
  	xlabel=Epochs,
	ylabel = $\%$ of caught balls,
	title = \texttt{Catch-v0},
	legend pos=south east,	
]

\addlegendentry{\texttt{Catch-v0}}
\addplot[blue, ultra thick] table[x=epochs,y=baseline] {./Results/logs/self_transfer_catch_v0.dat};
\addplot [name path=upper,draw=none, forget plot] table[x=epochs,y expr=\thisrow{baseline}+\thisrow{std_baseline}] {./Results/logs/self_transfer_catch_v0.dat};
\addplot [name path=lower,draw=none, forget plot] table[x=epochs,y expr=\thisrow{baseline}-\thisrow{std_baseline}] {./Results/logs/self_transfer_catch_v0.dat};
\addplot [fill=blue!10, forget plot] fill between[of=upper and lower];

\addlegendentry{Only-Head}
\addplot[dotted, ultra thick,blue] table[x=epochs,y=only_head] {./Results/logs/self_transfer_catch_v0.dat};
\addplot [name path=upper,draw=none, forget plot] table[x=epochs,y expr=\thisrow{only_head}+\thisrow{std_only_head}] {./Results/logs/self_transfer_catch_v0.dat};
\addplot [name path=lower,draw=none, forget plot] table[x=epochs,y expr=\thisrow{only_head}-\thisrow{std_only_head}] {./Results/logs/self_transfer_catch_v0.dat};
\addplot [fill=blue!10, forget plot] fill between[of=upper and lower];

\addlegendentry{Fine-Tuning}
\addplot[dashed, ultra thick,blue] table[x=epochs,y=self_transfer_full_fine_tuning] {./Results/logs/self_transfer_catch_v0.dat};
\addplot [name path=upper,draw=none, forget plot] table[x=epochs,y expr=\thisrow{self_transfer_full_fine_tuning}+\thisrow{std_self_transfer_full_fine_tuning}] {./Results/logs/self_transfer_catch_v0.dat};
\addplot [name path=lower,draw=none, forget plot] table[x=epochs,y expr=\thisrow{self_transfer_full_fine_tuning}-\thisrow{std_self_transfer_full_fine_tuning}] {./Results/logs/self_transfer_catch_v0.dat};
\addplot [fill=blue!10, forget plot] fill between[of=upper and lower];

\end{axis}

\begin{axis}
[	
	at={(ax1.south east)},
	xshift=1cm,	
	grid style={dashed,gray},
	grid = both, 
	tick style=black,
  	xlabel=Epochs,
	title = \texttt{Catch-v1},
	legend pos=south east,	
]

\addlegendentry{\texttt{Catch-v1}}
\addplot[yellow, ultra thick] table[x=epochs,y=baseline] {./Results/logs/self_transfer_catch_v1.dat};
\addplot [name path=upper,draw=none, forget plot] table[x=epochs,y expr=\thisrow{baseline}+\thisrow{std_baseline}] {./Results/logs/self_transfer_catch_v1.dat};
\addplot [name path=lower,draw=none, forget plot] table[x=epochs,y expr=\thisrow{baseline}-\thisrow{std_baseline}] {./Results/logs/self_transfer_catch_v1.dat};
\addplot [fill=yellow!10, forget plot] fill between[of=upper and lower];

\addlegendentry{Only-Head}
\addplot[dotted, ultra thick,yellow] table[x=epochs,y=only_head] {./Results/logs/self_transfer_catch_v1.dat};
\addplot [name path=upper,draw=none, forget plot] table[x=epochs,y expr=\thisrow{only_head}+\thisrow{std_only_head}] {./Results/logs/self_transfer_catch_v1.dat};
\addplot [name path=lower,draw=none, forget plot] table[x=epochs,y expr=\thisrow{only_head}-\thisrow{std_only_head}] {./Results/logs/self_transfer_catch_v1.dat};
\addplot [fill=yellow!10, forget plot] fill between[of=upper and lower];

\addlegendentry{Fine-Tuning}
\addplot[dashed, ultra thick,yellow] table[x=epochs,y=self_transfer_full_fine_tuning] {./Results/logs/self_transfer_catch_v1.dat};
\addplot [name path=upper,draw=none] table[x=epochs,y expr=\thisrow{self_transfer_full_fine_tuning}+\thisrow{std_self_transfer_full_fine_tuning}] {./Results/logs/self_transfer_catch_v1.dat};
\addplot [name path=lower,draw=none] table[x=epochs,y expr=\thisrow{self_transfer_full_fine_tuning}-\thisrow{std_self_transfer_full_fine_tuning}] {./Results/logs/self_transfer_catch_v1.dat};
\addplot [fill=yellow!10] fill between[of=upper and lower];

\end{axis}

\begin{axis}
[	
	at={(ax1.south east)},
	xshift=9cm,	
	grid style={dashed,gray},
	grid = both, 
	tick style=black,
  	xlabel=Epochs,
	title = \texttt{Catch-v2},
	legend pos=south east,	
]

\addlegendentry{\texttt{Catch-v2}}
\addplot[red, ultra thick] table[x=epochs,y=baseline] {./Results/logs/self_transfer_catch_v2.dat};
\addplot [name path=upper,draw=none, forget plot] table[x=epochs,y expr=\thisrow{baseline}+\thisrow{std_baseline}] {./Results/logs/self_transfer_catch_v2.dat};
\addplot [name path=lower,draw=none, forget plot] table[x=epochs,y expr=\thisrow{baseline}-\thisrow{std_baseline}] {./Results/logs/self_transfer_catch_v2.dat};
\addplot [fill=red!10, forget plot] fill between[of=upper and lower];

\addlegendentry{Only-Head}
\addplot[dotted, ultra thick,red] table[x=epochs,y=only_head] {./Results/logs/self_transfer_catch_v2.dat};
\addplot [name path=upper,draw=none, forget plot] table[x=epochs,y expr=\thisrow{only_head}+\thisrow{std_only_head}] {./Results/logs/self_transfer_catch_v2.dat};
\addplot [name path=lower,draw=none, forget plot] table[x=epochs,y expr=\thisrow{only_head}-\thisrow{std_only_head}] {./Results/logs/self_transfer_catch_v2.dat};
\addplot [fill=red!10, forget plot] fill between[of=upper and lower];

\addlegendentry{Fine-Tuning}
\addplot[dashed, ultra thick,red] table[x=epochs,y=self_transfer_full_fine_tuning] {./Results/logs/self_transfer_catch_v2.dat};
\addplot [name path=upper,draw=none, forget plot] table[x=epochs,y expr=\thisrow{self_transfer_full_fine_tuning}+\thisrow{std_self_transfer_full_fine_tuning}] {./Results/logs/self_transfer_catch_v2.dat};
\addplot [name path=lower,draw=none, forget plot] table[x=epochs,y expr=\thisrow{self_transfer_full_fine_tuning}-\thisrow{std_self_transfer_full_fine_tuning}] {./Results/logs/self_transfer_catch_v2.dat};
\addplot [fill=red!10, forget plot] fill between[of=upper and lower];

\end{axis}

\begin{axis}
[	
	at={(ax1.south east)},
	xshift=17cm,
	grid style={dashed,gray},
	grid = both, 
	tick style=black,
  	xlabel=Epochs,
	title = \texttt{Catch-v3},
	legend pos=south east,	
]

\addlegendentry{\texttt{Catch-v3}}
\addplot[green, ultra thick] table[x=epochs,y=baseline] {./Results/logs/self_transfer_catch_v3.dat};
\addplot [name path=upper,draw=none, forget plot] table[x=epochs,y expr=\thisrow{baseline}+\thisrow{std_baseline}] {./Results/logs/self_transfer_catch_v3.dat};
\addplot [name path=lower,draw=none, forget plot] table[x=epochs,y expr=\thisrow{baseline}-\thisrow{std_baseline}] {./Results/logs/self_transfer_catch_v3.dat};
\addplot [fill=green!10, forget plot] fill between[of=upper and lower];

\addlegendentry{Only-Head}
\addplot[dotted, ultra thick, green] table[x=epochs,y=only_head] {./Results/logs/self_transfer_catch_v3.dat};
\addplot [name path=upper,draw=none, forget plot] table[x=epochs,y expr=\thisrow{only_head}+\thisrow{std_only_head}] {./Results/logs/self_transfer_catch_v3.dat};
\addplot [name path=lower,draw=none, forget plot] table[x=epochs,y expr=\thisrow{only_head}-\thisrow{std_only_head}] {./Results/logs/self_transfer_catch_v3.dat};
\addplot [fill=green!10, forget plot] fill between[of=upper and lower];

\addlegendentry{Fine-Tuning}
\addplot[dashed, ultra thick, green] table[x=epochs,y=self_transfer_full_fine_tuning] {./Results/logs/self_transfer_catch_v3.dat};
\addplot [name path=upper,draw=none, forget plot] table[x=epochs,y expr=\thisrow{self_transfer_full_fine_tuning}+\thisrow{std_self_transfer_full_fine_tuning}] {./Results/logs/self_transfer_catch_v3.dat};
\addplot [name path=lower,draw=none, forget plot] table[x=epochs,y expr=\thisrow{self_transfer_full_fine_tuning}-\thisrow{std_self_transfer_full_fine_tuning}] {./Results/logs/self_transfer_catch_v3.dat};
\addplot [fill=green!10, forget plot] fill between[of=upper and lower];

\end{axis}

\end{tikzpicture}%
}
\caption{The results of our self-transfer experiments. From left to right the performance obtained on \texttt{Catchv-0}, \texttt{Catch-v1}, \texttt{Catch-v2} and \texttt{Catch-v3} after either training only the last linear layer of a pre-trained Deep-Q Network (dotted lines), or after wholly fine-tuning the model (dashed lines). We can see that the former transfer learning strategy yields significantly better results, and that a fine-tuning approach results in networks that in three cases out of four are not even able to transfer to themselves.}
\label{fig:catch_self_transfer}
\end{figure}
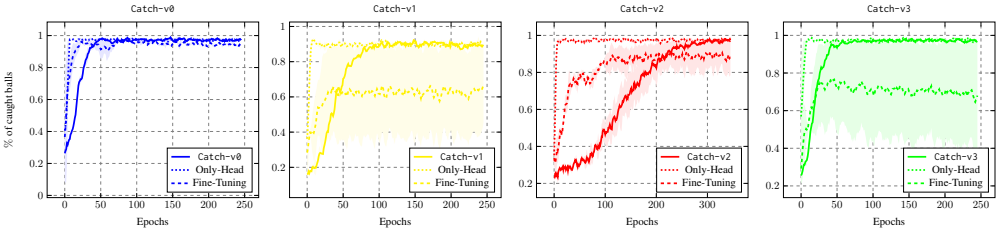

The surprisingly poor transfer learning performance observed in the previous experiment made us question the level of transferability of Deep-Q Networks even more. To further characterize their TL properties, we decided to investigate whether pre-trained DRL agents are at least able to transfer to themselves. To this end, we studied what happens when a DQN agent gets transferred to a version of \texttt{Catch} that matches with the version of the game that was also used during the pre-training stage. This experiment is in large part identical to the one presented in Sec. \ref{sec:from_one_catch_to_another}, with the only difference being that now $\mathcal{M}_S = \mathcal{M}_T$. Moreover, differently from the previous study, we now also investigate what happens if instead of fine-tuning the network completely, we just use the pre-trained DQN agent as a simple feature extractor, therefore only training its last layer (the head) responsible for estimating the different state-action values. 
\begin{wraptable}{r}{9cm}
	\centering
	\caption{The area ratio scores obtained after performing self-transfer. We can see that if only the last linear layer is trained, then positive transfer is obtained on all \texttt{Catch} environments, whereas if the network is fine-tuned, positive transfer is (in part) only obtained on \texttt{Catch-v2}.}

\newcommand*{\MinNumber}{-0.674}%
\newcommand*{\MidNumber}{0.0}%
\newcommand*{\MaxNumber}{0.674}%

\newcommand{\ApplyGradient}[1]{%
  \IfDecimal{#1}{
    \ifdim #1 pt > \MidNumber pt%
    \pgfmathsetmacro{\PercentColor}{max(min(100.0*(#1-\MidNumber)/(\MaxNumber-\MidNumber),100.0),0.0)}%
    \edef\x{\noexpand\cellcolor{\colorpos!\PercentColor!white}}\x\textcolor{black}{#1}%
    \else%
    \pgfmathsetmacro{\PercentColor}{max(min(100.0*(\MidNumber-#1)/(\MidNumber-\MinNumber),100.0),0.0)}%
    \edef\x{\noexpand\cellcolor{\colorneg!\PercentColor!white}}\x\textcolor{black}{#1}%
    \fi%
    }{#1}
}

\newcolumntype{R}{>{\collectcell\ApplyGradient}{c}<{\endcollectcell}}
\begin{tabular}{c|R|R|R|R}
\hline 
           &  Catch-v0 & Catch-v1 & Catch-v2 & Catch-v3 \\ \hline \hline
Only-Head   & 0.05     & 0.141    & 0.674    & 0.059    \\
Fine-Tuning & 0.017    & -0.218   & 0.393    & -0.236   \\   
\end{tabular}	

	\label{tab:self_tl_area_ratio}
\end{wraptable}
Our results are presented in Fig. \ref{fig:catch_self_transfer}, where for each \texttt{Catch} version, the full lines represent the performance of a network that is trained from scratch, whereas the dashed and dotted lines respectively report the performance that is obtained when the network is either used as simple feature extractor or entirely fine-tuned. We can see that if a pre-trained Deep-Q Network is used as a simple feature extractor, the agent can converge to the optimal policy almost immediately. In fact, as can consistently be observed in all plots of Fig. \ref{fig:catch_self_transfer} and from the results presented in Table \ref{tab:self_tl_area_ratio}, training only the last layer of a pre-trained network yields positive transfer for all different \texttt{Catch} versions. However, when a fine-tuning training strategy is adopted, much more surprising results have been obtained. First, and more importantly, we can see that despite all models showing some learning speed improvements at early training iterations, their final performance is never on par with the one that is obtained when the same kind of model is either used as a feature extractor or trained from scratch (the dotted lines are consistently below the dashed and full lines). While when it comes to \texttt{Catch-v0} the policy learned by a fine-tuned model still allows the agent to successfully catch $\approx 95\%$ of the falling balls, the same cannot be said when the models are tested on \texttt{Catch-v1, Catch-v2} and \texttt{Catch-v3}, where the difference in terms of performance between a model trained from scratch and a fine-tuned one is much more significant. Please also note that special attention should be given to \texttt{Catch-v2}, which is an environment where the area ratio score reported in Table \ref{tab:self_tl_area_ratio} can be misleading as it does not entirely reflect the quality of the final policy learned by the agent. In fact, while it is true that an $r$ value of $0.393$ is obtained, it is worth noting that a fine-tuned network converges to a policy that is significantly worse than the one of a network trained from scratch, as the agent is only able of catching $\approx 80\%$ of the falling balls. Second, as highlighted by the large variance across different training runs, fine-tuning on \texttt{Catch-v1} and \texttt{Catch-v3} resulted in highly unstable learning as well.    

\section{The Two Learning Phases of Deep-Q Networks}
\label{sec:hybrid_self_transfer}

\begin{figure}[ht]
\begin{minipage}{0.5\textwidth}
	\centering
\includegraphics[width=6.5cm]{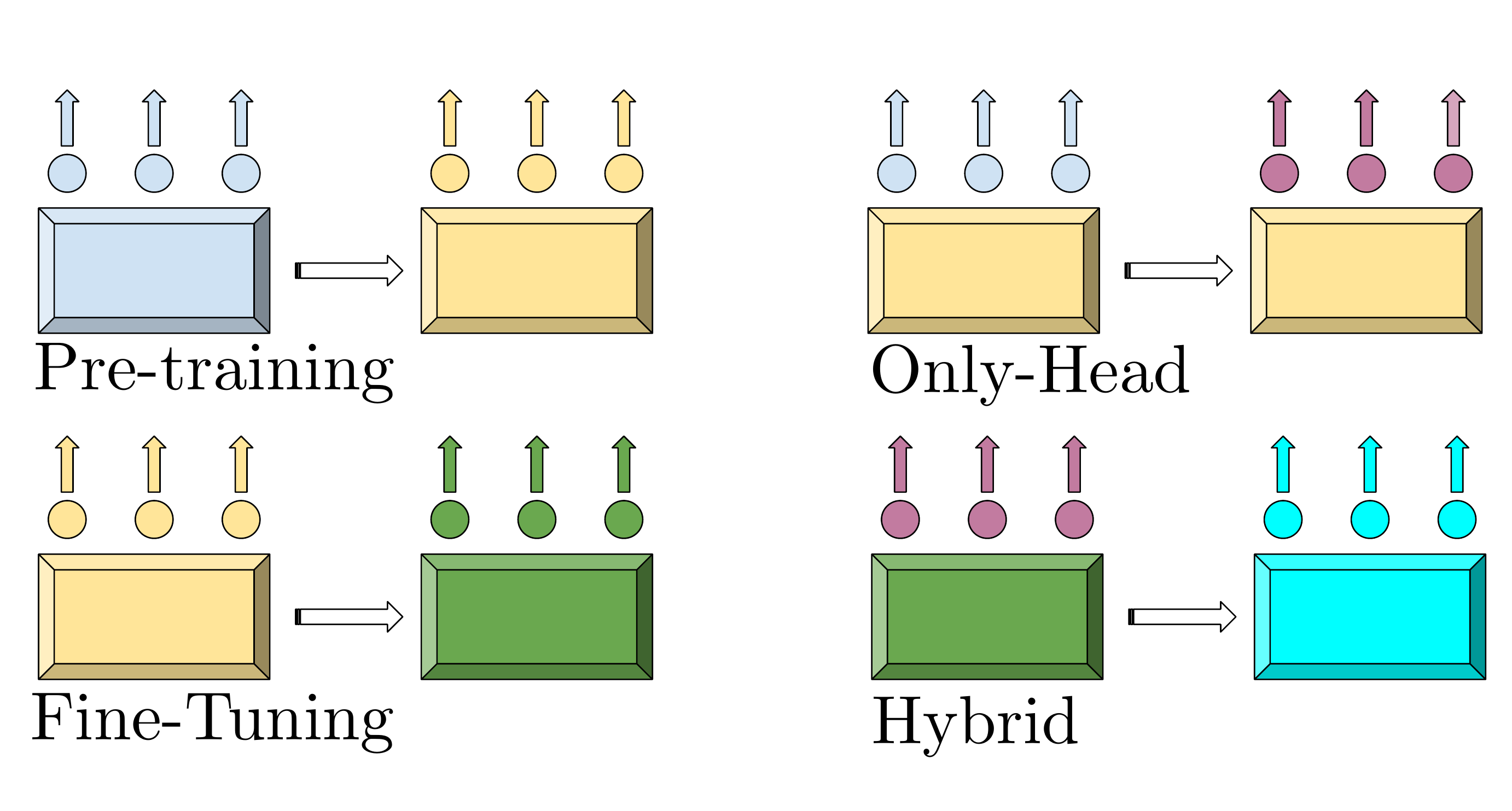}
\end{minipage}%
\begin{minipage}{0.5\textwidth}
	\centering
	\begin{tikzpicture}[scale=0.65]
\begin{axis}
[	grid style={dashed,gray},
	grid = both, 
	tick style=black,
  	xlabel=Epochs,
	ylabel = $\%$ of caught balls,
	legend pos=outer north east,
]

\addlegendentry{\texttt{Catch-v0}}

\addplot[blue, ultra thick] table[x=epochs,y=tl] {./Results/logs/final_tl_exp.dat};
\addplot[name path=upper,draw=none, forget plot] table[x=epochs,y expr=\thisrow{tl}+\thisrow{std_tl}] {./Results/logs/final_tl_exp.dat};
\addplot[name path=lower,draw=none, forget plot] table[x=epochs,y expr=\thisrow{tl}-\thisrow{std_tl}] {./Results/logs/final_tl_exp.dat};
\addplot[fill=blue!10, forget plot] fill between[of=upper and lower];

\addlegendentry{\texttt{Catch-v4}}
\addplot[ultra thick, purple] table[x=epochs,y=baseline] {./Results/logs/final_tl_exp.dat};
\addplot [name path=upper,draw=none, forget plot] table[x=epochs,y expr=\thisrow{baseline}+\thisrow{std_baseline}] {./Results/logs/final_tl_exp.dat};
\addplot [name path=lower,draw=none, forget plot] table[x=epochs,y expr=\thisrow{baseline}-\thisrow{std_baseline}] {./Results/logs/final_tl_exp.dat};
\addplot [fill=purple!10, forget plot] fill between[of=upper and lower];

\end{axis}
\end{tikzpicture}
\end{minipage}
\caption{Image on the left: a visualization of differently initialized Deep-Q Networks before and after training. Image on the right: a successful example of positive transfer.}
\label{fig:networks}
\end{figure}
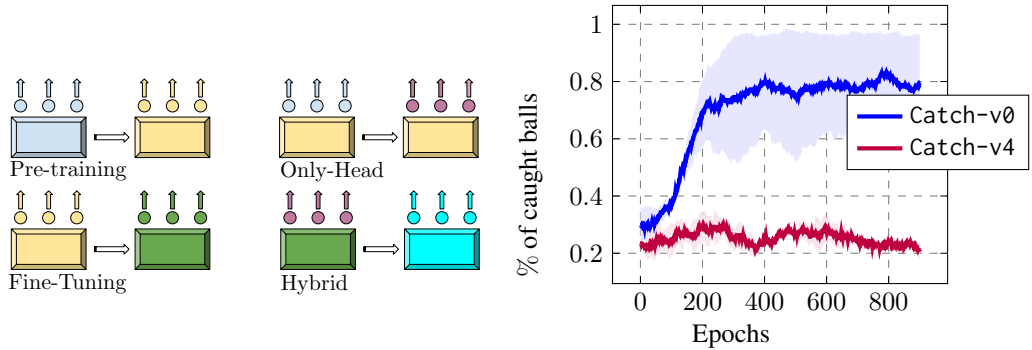

A prototypical Deep-Q Network takes as input an image representing the state of the environment and processes it through a series of convolutions and a fully connected layer. When this is done, it outputs as many $Q(s,a)$ values as there are actions available to the agent, a process that corresponds to learning a linear policy in the latent feature space of the network. It follows that by the end of training, such a model has to serve two purposes: it has to perform as a feature extractor as well as an optimal value function approximator.
Extracting relevant features from high dimensional inputs and learning an optimal value function can arguably be seen as two separate tasks; yet, despite their dissimilarity, we believe that they are more interconnected than one might expect. Specifically, we hypothesize that while learning, a Deep-Q Network has to carefully find a balance between training the parts that serve as feature extractors and the components that are responsible for estimating a policy. The poor TL performance observed throughout this work could therefore be the result of using models where the feature extractor component of an agent, as it comes as pre-trained, is too detached from the respective final layer of the network, which is randomly initialized instead.   
To show that a Deep-Q Network has to carefully coordinate training its feature extractor components and its final layer, let us consider the left image of Fig. \ref{fig:networks}. The figure depicts how the weights of an agent, whose feature extractor layers are represented by a square and the linear layer is represented through circles, change according to the self-TL experiments presented in Sec. \ref{sec:self_transfer}. Each experiment is represented through two networks, one on the left side of the arrow representing the source model, and a second one, on the right part of the arrow, obtained by the end of training. From top to bottom, and from left to right, the first three network pairs represent the training process of: a randomly initialized model trained from scratch, a pre-trained model whose only last linear layer is trained after random initialization, and a pre-trained agent who gets fully fine-tuned. Following the results presented in Sec. \ref{sec:self_transfer}, we know that positive transfer is only obtained when the last linear layer is trained in isolation after being randomly re-initialized, whereas negative transfer is obtained if a fine-tuning training strategy is adopted. 

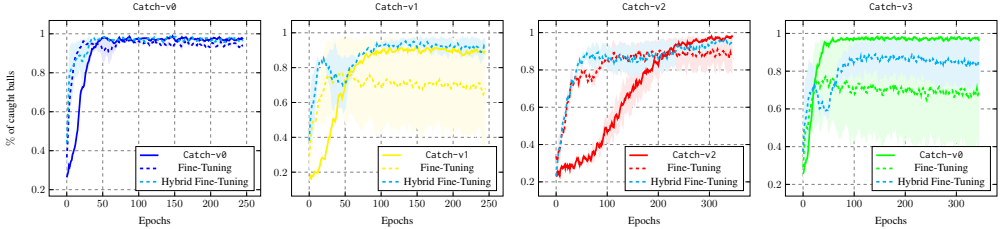
\begin{figure}[ht!]
\resizebox{0.45\textwidth}{!}{
\begin{tikzpicture}
\begin{axis}
[	
	name=ax1,	
	grid style={dashed,gray},
	grid = both, 
	tick style=black,
  	xlabel=Epochs,
	ylabel = $\%$ of caught balls,
	title = \texttt{Catch-v0},
	legend pos=south east,	
]

\addlegendentry{\texttt{Catch-v0}}
\addplot[blue, ultra thick] table[x=epochs,y=baseline] {./Results/logs/hybrid_self_tl_v0.dat};
\addplot [name path=upper,draw=none, forget plot] table[x=epochs,y expr=\thisrow{baseline}+\thisrow{std_baseline}] {./Results/logs/hybrid_self_tl_v0.dat};
\addplot [name path=lower,draw=none, forget plot] table[x=epochs,y expr=\thisrow{baseline}-\thisrow{std_baseline}] {./Results/logs/hybrid_self_tl_v0.dat};
\addplot [fill=blue!10, forget plot] fill between[of=upper and lower];

\addlegendentry{Fine-Tuning}
\addplot[dashed, ultra thick,blue] table[x=epochs,y=self_transfer_full_fine_tuning] {./Results/logs/hybrid_self_tl_v0.dat};
\addplot [name path=upper,draw=none, forget plot] table[x=epochs,y expr=\thisrow{self_transfer_full_fine_tuning}+\thisrow{std_self_transfer_full_fine_tuning}] {./Results/logs/hybrid_self_tl_v0.dat};
\addplot [name path=lower,draw=none, forget plot] table[x=epochs,y expr=\thisrow{self_transfer_full_fine_tuning}-\thisrow{std_self_transfer_full_fine_tuning}] {./Results/logs/hybrid_self_tl_v0.dat};
\addplot [fill=blue!10, forget plot] fill between[of=upper and lower];

\addlegendentry{Hybrid Fine-Tuning}
\addplot[dashed, ultra thick,cyan] table[x=epochs,y=hybrid_self_transfer] {./Results/logs/hybrid_self_tl_v0.dat};
\addplot [name path=upper,draw=none, forget plot] table[x=epochs,y expr=\thisrow{hybrid_self_transfer}+\thisrow{std_hybrid_self_transfer}] {./Results/logs/hybrid_self_tl_v0.dat};
\addplot [name path=lower,draw=none, forget plot] table[x=epochs,y expr=\thisrow{hybrid_self_transfer}-\thisrow{std_hybrid_self_transfer}] {./Results/logs/hybrid_self_tl_v0.dat};
\addplot [fill=cyan!10, forget plot] fill between[of=upper and lower];

\end{axis}

\begin{axis}
[	
	at={(ax1.south east)},
	xshift=1cm,	
	grid style={dashed,gray},
	grid = both, 
	tick style=black,
  	xlabel=Epochs,
	title = \texttt{Catch-v1},
	legend pos=south east,	
]

\addlegendentry{\texttt{Catch-v1}}
\addplot[yellow, ultra thick] table[x=epochs,y=baseline] {./Results/logs/self_transfer_catch_v1.dat};
\addplot [name path=upper,draw=none, forget plot] table[x=epochs,y expr=\thisrow{baseline}+\thisrow{std_baseline}] {./Results/logs/self_transfer_catch_v1.dat};
\addplot [name path=lower,draw=none, forget plot] table[x=epochs,y expr=\thisrow{baseline}-\thisrow{std_baseline}] {./Results/logs/self_transfer_catch_v1.dat};
\addplot [fill=yellow!10, forget plot] fill between[of=upper and lower];

\addlegendentry{Fine-Tuning}
\addplot[dashed, ultra thick,yellow] table[x=epochs,y=self_transfer_full_fine_tuning] {./Results/logs/hybrid_self_tl_v1.dat};
\addplot [name path=upper,draw=none, forget plot] table[x=epochs,y expr=\thisrow{self_transfer_full_fine_tuning}+\thisrow{std_self_transfer_full_fine_tuning}] {./Results/logs/hybrid_self_tl_v1.dat};
\addplot [name path=lower,draw=none, forget plot] table[x=epochs,y expr=\thisrow{self_transfer_full_fine_tuning}-\thisrow{std_self_transfer_full_fine_tuning}] {./Results/logs/hybrid_self_tl_v1.dat};
\addplot [fill=yellow!10, forget plot] fill between[of=upper and lower];

\addlegendentry{Hybrid Fine-Tuning}
\addplot[dashed, ultra thick,cyan] table[x=epochs,y=hybrid_self_transfer] {./Results/logs/hybrid_self_tl_v1.dat};
\addplot [name path=upper,draw=none, forget plot] table[x=epochs,y expr=\thisrow{hybrid_self_transfer}+\thisrow{std_hybrid_self_transfer}] {./Results/logs/hybrid_self_tl_v1.dat};
\addplot [name path=lower,draw=none, forget plot] table[x=epochs,y expr=\thisrow{hybrid_self_transfer}-\thisrow{std_hybrid_self_transfer}] {./Results/logs/hybrid_self_tl_v1.dat};
\addplot [fill=cyan!10, forget plot] fill between[of=upper and lower];

\end{axis}

\begin{axis}
[	
	at={(ax1.south east)},
	xshift=9cm,	
	grid style={dashed,gray},
	grid = both, 
	tick style=black,
  	xlabel=Epochs,
	title = \texttt{Catch-v2},
	legend pos=south east,	
]

\addlegendentry{\texttt{Catch-v2}}
\addplot[red, ultra thick] table[x=epochs,y=baseline] {./Results/logs/hybrid_self_tl_v2.dat};
\addplot [name path=upper,draw=none, forget plot] table[x=epochs,y expr=\thisrow{baseline}+\thisrow{std_baseline}] {./Results/logs/hybrid_self_tl_v2.dat};
\addplot [name path=lower,draw=none, forget plot] table[x=epochs,y expr=\thisrow{baseline}-\thisrow{std_baseline}] {./Results/logs/hybrid_self_tl_v2.dat};
\addplot [fill=red!10, forget plot] fill between[of=upper and lower];

\addlegendentry{Fine-Tuning}
\addplot[dashed, ultra thick,red] table[x=epochs,y=self_transfer_full_fine_tuning] {./Results/logs/hybrid_self_tl_v2.dat};
\addplot [name path=upper,draw=none, forget plot] table[x=epochs,y expr=\thisrow{self_transfer_full_fine_tuning}+\thisrow{std_self_transfer_full_fine_tuning}] {./Results/logs/hybrid_self_tl_v2.dat};
\addplot [name path=lower,draw=none, forget plot] table[x=epochs,y expr=\thisrow{self_transfer_full_fine_tuning}-\thisrow{std_self_transfer_full_fine_tuning}] {./Results/logs/hybrid_self_tl_v2.dat};
\addplot [fill=red!10, forget plot] fill between[of=upper and lower];

\addlegendentry{Hybrid Fine-Tuning}
\addplot[dashed, ultra thick,cyan] table[x=epochs,y=hybrid_self_transfer] {./Results/logs/hybrid_self_tl_v2.dat};
\addplot [name path=upper,draw=none, forget plot] table[x=epochs,y expr=\thisrow{hybrid_self_transfer}+\thisrow{std_hybrid_self_transfer}] {./Results/logs/hybrid_self_tl_v2.dat};
\addplot [name path=lower,draw=none, forget plot] table[x=epochs,y expr=\thisrow{hybrid_self_transfer}-\thisrow{std_hybrid_self_transfer}] {./Results/logs/hybrid_self_tl_v2.dat};
\addplot [fill=cyan!10, forget plot] fill between[of=upper and lower];

\end{axis}

\begin{axis}
[	
	at={(ax1.south east)},
	xshift=17cm,	
	grid style={dashed,gray},
	grid = both, 
	tick style=black,
  	xlabel=Epochs,
	title = \texttt{Catch-v3},
	legend pos=south east,	
]

\addlegendentry{\texttt{Catch-v0}}
\addplot[green, ultra thick] table[x=epochs,y=baseline] {./Results/logs/hybrid_self_tl_v3.dat};
\addplot [name path=upper,draw=none, forget plot] table[x=epochs,y expr=\thisrow{baseline}+\thisrow{std_baseline}] {./Results/logs/hybrid_self_tl_v3.dat};
\addplot [name path=lower,draw=none, forget plot] table[x=epochs,y expr=\thisrow{baseline}-\thisrow{std_baseline}] {./Results/logs/hybrid_self_tl_v3.dat};
\addplot [fill=green!10, forget plot] fill between[of=upper and lower];

\addlegendentry{Fine-Tuning}
\addplot[dashed, ultra thick,green] table[x=epochs,y=self_transfer_full_fine_tuning] {./Results/logs/hybrid_self_tl_v3.dat};
\addplot [name path=upper,draw=none, forget plot] table[x=epochs,y expr=\thisrow{self_transfer_full_fine_tuning}+\thisrow{std_self_transfer_full_fine_tuning}] {./Results/logs/hybrid_self_tl_v3.dat};
\addplot [name path=lower,draw=none, forget plot] table[x=epochs,y expr=\thisrow{self_transfer_full_fine_tuning}-\thisrow{std_self_transfer_full_fine_tuning}] {./Results/logs/hybrid_self_tl_v3.dat};
\addplot [fill=green!10, forget plot] fill between[of=upper and lower];

\addlegendentry{Hybrid Fine-Tuning}
\addplot[dashed, ultra thick,cyan] table[x=epochs,y=hybrid_self_transfer] {./Results/logs/hybrid_self_tl_v3.dat};
\addplot [name path=upper,draw=none, forget plot] table[x=epochs,y expr=\thisrow{hybrid_self_transfer}+\thisrow{std_hybrid_self_transfer}] {./Results/logs/hybrid_self_tl_v3.dat};
\addplot [name path=lower,draw=none, forget plot] table[x=epochs,y expr=\thisrow{hybrid_self_transfer}-\thisrow{std_hybrid_self_transfer}] {./Results/logs/hybrid_self_tl_v3.dat};
\addplot [fill=cyan!10, forget plot] fill between[of=upper and lower];

\end{axis}

\end{tikzpicture}%
}
\caption{The performance (in cyan) of a fine-tuned pre-trained network whose last layer is initialized with parameters that yielded positive transfer, whereas its convolutional and fully conntected layers are initialized with parameters that yielded negative transfer.}
\label{fig:hybrid_self_transfer}
\end{figure}

We now investigate the TL performance of a model that is a combination of the Only-Head and Fine-Tuning settings (see bottom right image of Fig. \ref{fig:networks} for a visualization). Specifically, we fine-tune a Deep-Q Network whose last linear layer is initialized with the parameters that yielded positive transfer in Sec. \ref{sec:self_transfer} (Only-Head in Fig. \ref{fig:networks}), whereas its convolutional and fully connected layers are initialized with the parameters that yielded negative transfer (Fine-Tuning in Fig. \ref{fig:networks}). We visualize the self-transfer performance of these models, denoted as `Hybrid", as they are a hybrid combination of two differently pre-trained networks, in Fig. \ref{fig:hybrid_self_transfer} with a cyan dashed dotted line. We can observe that learning is characterized by a very atypical behavior: the network starts by improving its performance (thanks to the already trained final layer); it then goes through a second stage where it starts to perform more poorly (due to the poor feature extractor part), and then finally starts learning stably (when the feature extractor and the head of the model are synchronized). We believe that the poor TL performance observed throughout this work is, therefore, the result of models which could not find a balance between a randomly initialized head and their respective pre-trained layers which are too biased towards the source task.   

Based on these results, one critical question still remains to be answered: how come positive transfer for some Atari games was observed in Sec. \ref{sec:empirical_study}?. We believe that the answer to this question does not lie within the feature representations that a Deep-Q Network learns, but rather in some inner properties of the environment that is used as target task and that favors a Deep-Q Network to correctly synchronize its components. As a proof of concept, we have created one final \texttt{Catch} environment, called \texttt{Catch-v4}, which is identical to \texttt{Catch-v0} with the only difference being that a positive reward is returned to the agent only if it manages to catch five falling balls in a row. We can see from the right image of Fig. \ref{fig:networks} that a model trained from scratch (represented by the purple line) is not able to improve its policy over time at all, as the reward signal is probably too sparse for learning, whereas a \texttt{Catch-v0} model is now able to yield positive transfer. We hence believe that some environments, if combined with certain learning algorithms, are more prone to positive transfer than others, as was, e.g., the case for \texttt{Fishing Derby} and DQV-Learning where positive transfer was observed no matter what source task was used for pre-training. This is not due to the representations that are learned by a pre-trained network but rather because of some specific dynamics within the target MDP $\mathcal{M}_T$. 

\section{Related Work \& Conclusion}

The closest research to the one presented in this work is certainly that of \cite{farebrother2018generalization} and \cite{tyo2020transferable}, who also studied the generalization properties of Deep-Q Networks in a model-free DRL context. Our extensive experiments confirm some of the preliminary claims that were made by the former about the potentially poor TL properties of Deep-Q Networks, but contradict the study of the latter, who suggested that fine-tuning DRL agents results in positive transfer when moving from a simpler task to a harder. While both works are certainly valuable, we also believe that their experimental results are not as thorough and on par with the ones of this study as they both considered a very limited number of RL problems (four and three respectively), therefore leaving the question \textit{"How transferable are Deep-Q Networks?"} unanswered. We believe that the answer to it is \textit{"barely"}, but we also believe that their poor TL properties, as well as the learning dynamics identified in Sec. \ref{sec:hybrid_self_transfer}, are inherent to this family of algorithms only. In fact, it is worth noting that several works describing the benefits of TL in RL do exist (but they all differ from the study presented in this work): \cite{tirinzoni2018transfer} show that it possible to successfully transfer value functions across tasks, yet their work does not consider deep networks as function approximators but rather Gaussian mixtures. \cite{parisotto2015actor} show that it can be beneficial to fine-tune a pre-trained DRL agent, but they consider multi-task learning and policy gradient algorithms as a way of pre-training. \cite{rusu2016progressive} also show that fine-tuning can be beneficial, but in the context of progressive networks and again of policy gradient techniques. Similar conclusions for Actor-Critic algorithms can also be found in the works of \cite{zhu2017target} and \cite{chen2021improving}. Furthermore, \cite{landolfi2019model} and \cite{sasso2021fractional} show that fine-tuning a pre-trained network can be beneficial for DRL tasks, but for model-based RL approaches, which are again part of a family of techniques that is different from the ones analyzed in this work. To conclude, we would like to stress out that despite the overall poor TL performance observed throughout this paper, positive transfer can nevertheless be obtained in a model-free DRL setup, and hope that this paper can serve as a solid starting point for the DRL community which is interested in designing general and transferable agents.  

\section*{Acknowledgements}
Matthia Sabatelli acknowledges the financial support of BELSPO, Federal Public Planning Service
Science Policy, Belgium, in the context of the BRAIN-be project. He also wishes to thank Gilles
Louppe and Antoine Wehenkel for their valuable feedback on preliminary versions of this work.
Lastly, one final thought goes to Marco A. Wiering, his teacher; supervisor; colleague; but mostly
beloved friend: \textit{‘I will miss you deeply”}.

\bibliographystyle{abbrvnat}
\bibliography{my_bib.bib}

\begin{thebibliography}{50}
\providecommand{\natexlab}[1]{#1}
\providecommand{\url}[1]{\texttt{#1}}
\expandafter\ifx\csname urlstyle\endcsname\relax
  \providecommand{\doi}[1]{doi: #1}\else
  \providecommand{\doi}{doi: \begingroup \urlstyle{rm}\Url}\fi

\bibitem[Aittahar et~al.(2020)Aittahar, Fonteneau, and
  Ernst]{aittahar2020empirical}
S.~Aittahar, R.~Fonteneau, and D.~Ernst.
\newblock Empirical analysis of policy gradient algorithms where starting
  states are sampled accordingly to most frequently visited states.
\newblock \emph{IFAC-PapersOnLine}, 53\penalty0 (2):\penalty0 8097--8104, 2020.

\bibitem[Bellemare et~al.(2013)Bellemare, Naddaf, Veness, and
  Bowling]{bellemare2013arcade}
M.~G. Bellemare, Y.~Naddaf, J.~Veness, and M.~Bowling.
\newblock The arcade learning environment: An evaluation platform for general
  agents.
\newblock \emph{Journal of Artificial Intelligence Research}, 47:\penalty0
  253--279, 2013.

\bibitem[Chen et~al.(2021)Chen, Lee, Srinivas, and Abbeel]{chen2021improving}
L.~Chen, K.~Lee, A.~Srinivas, and P.~Abbeel.
\newblock Improving computational efficiency in visual reinforcement learning
  via stored embeddings.
\newblock \emph{arXiv preprint arXiv:2103.02886}, 2021.

\bibitem[Dom{\'\i}nguez~S{\'a}nchez et~al.(2019)Dom{\'\i}nguez~S{\'a}nchez,
  Huertas-Company, Bernardi, Kaviraj, Fischer, Abbott, Abdalla, Annis, Avila,
  Brooks, et~al.]{dominguez2019transfer}
H.~Dom{\'\i}nguez~S{\'a}nchez, M.~Huertas-Company, M.~Bernardi, S.~Kaviraj,
  J.~Fischer, T.~Abbott, F.~Abdalla, J.~Annis, S.~Avila, D.~Brooks, et~al.
\newblock Transfer learning for galaxy morphology from one survey to another.
\newblock \emph{Monthly Notices of the Royal Astronomical Society},
  484\penalty0 (1):\penalty0 93--100, 2019.

\bibitem[Farebrother et~al.(2018)Farebrother, Machado, and
  Bowling]{farebrother2018generalization}
J.~Farebrother, M.~C. Machado, and M.~Bowling.
\newblock Generalization and regularization in dqn.
\newblock \emph{arXiv preprint arXiv:1810.00123}, 2018.

\bibitem[Fran{\c{c}}ois-Lavet et~al.(2018)Fran{\c{c}}ois-Lavet, Henderson,
  Islam, Bellemare, and Pineau]{franccois2018introduction}
V.~Fran{\c{c}}ois-Lavet, P.~Henderson, R.~Islam, M.~G. Bellemare, and
  J.~Pineau.
\newblock An introduction to deep reinforcement learning.
\newblock \emph{arXiv preprint arXiv:1811.12560}, 2018.

\bibitem[Fujimoto et~al.(2018)Fujimoto, Hoof, and
  Meger]{fujimoto2018addressing}
S.~Fujimoto, H.~Hoof, and D.~Meger.
\newblock Addressing function approximation error in actor-critic methods.
\newblock In \emph{International Conference on Machine Learning}, pages
  1587--1596. PMLR, 2018.

\bibitem[Ha and Schmidhuber(2018)]{ha2018world}
D.~Ha and J.~Schmidhuber.
\newblock World models.
\newblock \emph{arXiv preprint arXiv:1803.10122}, 2018.

\bibitem[Haarnoja et~al.(2018)Haarnoja, Zhou, Abbeel, and
  Levine]{haarnoja2018soft}
T.~Haarnoja, A.~Zhou, P.~Abbeel, and S.~Levine.
\newblock Soft actor-critic: Off-policy maximum entropy deep reinforcement
  learning with a stochastic actor.
\newblock In \emph{International Conference on Machine Learning}, pages
  1861--1870. PMLR, 2018.

\bibitem[Hafner et~al.(2019{\natexlab{a}})Hafner, Lillicrap, Ba, and
  Norouzi]{hafner2019dream}
D.~Hafner, T.~Lillicrap, J.~Ba, and M.~Norouzi.
\newblock Dream to control: Learning behaviors by latent imagination.
\newblock \emph{arXiv preprint arXiv:1912.01603}, 2019{\natexlab{a}}.

\bibitem[Hafner et~al.(2019{\natexlab{b}})Hafner, Lillicrap, Fischer, Villegas,
  Ha, Lee, and Davidson]{hafner2019learning}
D.~Hafner, T.~Lillicrap, I.~Fischer, R.~Villegas, D.~Ha, H.~Lee, and
  J.~Davidson.
\newblock Learning latent dynamics for planning from pixels.
\newblock In \emph{International Conference on Machine Learning}, pages
  2555--2565. PMLR, 2019{\natexlab{b}}.

\bibitem[Hafner et~al.(2020)Hafner, Lillicrap, Norouzi, and
  Ba]{hafner2020mastering}
D.~Hafner, T.~Lillicrap, M.~Norouzi, and J.~Ba.
\newblock Mastering atari with discrete world models.
\newblock \emph{arXiv preprint arXiv:2010.02193}, 2020.

\bibitem[Henderson et~al.(2018)Henderson, Islam, Bachman, Pineau, Precup, and
  Meger]{henderson2018deep}
P.~Henderson, R.~Islam, P.~Bachman, J.~Pineau, D.~Precup, and D.~Meger.
\newblock Deep reinforcement learning that matters.
\newblock In \emph{Thirty-Second AAAI Conference on Artificial Intelligence},
  2018.

\bibitem[Ho and Kim(2021)]{ho2021evaluation}
N.~Ho and Y.-C. Kim.
\newblock Evaluation of transfer learning in deep convolutional neural network
  models for cardiac short axis slice classification.
\newblock \emph{Scientific reports}, 11\penalty0 (1):\penalty0 1--11, 2021.

\bibitem[Huh et~al.(2016)Huh, Agrawal, and Efros]{huh2016makes}
M.~Huh, P.~Agrawal, and A.~A. Efros.
\newblock What makes imagenet good for transfer learning?
\newblock \emph{arXiv preprint arXiv:1608.08614}, 2016.

\bibitem[Kaiser et~al.(2019)Kaiser, Babaeizadeh, Milos, Osinski, Campbell,
  Czechowski, Erhan, Finn, Kozakowski, Levine, et~al.]{kaiser2019model}
L.~Kaiser, M.~Babaeizadeh, P.~Milos, B.~Osinski, R.~H. Campbell, K.~Czechowski,
  D.~Erhan, C.~Finn, P.~Kozakowski, S.~Levine, et~al.
\newblock Model-based reinforcement learning for atari.
\newblock \emph{arXiv preprint arXiv:1903.00374}, 2019.

\bibitem[Landolfi et~al.(2019)Landolfi, Thomas, and Ma]{landolfi2019model}
N.~C. Landolfi, G.~Thomas, and T.~Ma.
\newblock A model-based approach for sample-efficient multi-task reinforcement
  learning.
\newblock \emph{arXiv preprint arXiv:1907.04964}, 2019.

\bibitem[Lazaric(2012)]{lazaric2012transfer}
A.~Lazaric.
\newblock Transfer in reinforcement learning: a framework and a survey.
\newblock In \emph{Reinforcement Learning}, pages 143--173. Springer, 2012.

\bibitem[Lillicrap et~al.(2015)Lillicrap, Hunt, Pritzel, Heess, Erez, Tassa,
  Silver, and Wierstra]{lillicrap2015continuous}
T.~P. Lillicrap, J.~J. Hunt, A.~Pritzel, N.~Heess, T.~Erez, Y.~Tassa,
  D.~Silver, and D.~Wierstra.
\newblock Continuous control with deep reinforcement learning.
\newblock \emph{arXiv preprint arXiv:1509.02971}, 2015.

\bibitem[Mensink et~al.(2021)Mensink, Uijlings, Kuznetsova, Gygli, and
  Ferrari]{mensink2021factors}
T.~Mensink, J.~Uijlings, A.~Kuznetsova, M.~Gygli, and V.~Ferrari.
\newblock Factors of influence for transfer learning across diverse appearance
  domains and task types.
\newblock \emph{arXiv preprint arXiv:2103.13318}, 2021.

\bibitem[Mnih et~al.(2013)Mnih, Kavukcuoglu, Silver, Graves, Antonoglou,
  Wierstra, and Riedmiller]{mnih2013playing}
V.~Mnih, K.~Kavukcuoglu, D.~Silver, A.~Graves, I.~Antonoglou, D.~Wierstra, and
  M.~Riedmiller.
\newblock Playing atari with deep reinforcement learning.
\newblock \emph{arXiv preprint arXiv:1312.5602}, 2013.

\bibitem[Mnih et~al.(2014)Mnih, Heess, Graves, et~al.]{mnih2014recurrent}
V.~Mnih, N.~Heess, A.~Graves, et~al.
\newblock Recurrent models of visual attention.
\newblock In \emph{Advances in neural information processing systems}, pages
  2204--2212, 2014.

\bibitem[Mnih et~al.(2015)Mnih, Kavukcuoglu, Silver, Rusu, Veness, Bellemare,
  Graves, Riedmiller, Fidjeland, Ostrovski, et~al.]{mnih2015human}
V.~Mnih, K.~Kavukcuoglu, D.~Silver, A.~A. Rusu, J.~Veness, M.~G. Bellemare,
  A.~Graves, M.~Riedmiller, A.~K. Fidjeland, G.~Ostrovski, et~al.
\newblock Human-level control through deep reinforcement learning.
\newblock \emph{Nature}, 518\penalty0 (7540):\penalty0 529, 2015.

\bibitem[Mnih et~al.(2016)Mnih, Badia, Mirza, Graves, Lillicrap, Harley,
  Silver, and Kavukcuoglu]{mnih2016asynchronous}
V.~Mnih, A.~P. Badia, M.~Mirza, A.~Graves, T.~Lillicrap, T.~Harley, D.~Silver,
  and K.~Kavukcuoglu.
\newblock Asynchronous methods for deep reinforcement learning.
\newblock In \emph{International conference on machine learning}, pages
  1928--1937, 2016.

\bibitem[Mormont et~al.(2018)Mormont, Geurts, and
  Mar{\'e}e]{mormont2018comparison}
R.~Mormont, P.~Geurts, and R.~Mar{\'e}e.
\newblock Comparison of deep transfer learning strategies for digital
  pathology.
\newblock In \emph{Proceedings of the IEEE Conference on Computer Vision and
  Pattern Recognition Workshops}, pages 2262--2271, 2018.

\bibitem[Obando-Ceron and Castro(2020)]{obando2020revisiting}
J.~S. Obando-Ceron and P.~S. Castro.
\newblock Revisiting rainbow: Promoting more insightful and inclusive deep
  reinforcement learning research.
\newblock \emph{arXiv preprint arXiv:2011.14826}, 2020.

\bibitem[Pan and Yang(2009)]{pan2009survey}
S.~J. Pan and Q.~Yang.
\newblock A survey on transfer learning.
\newblock \emph{IEEE Transactions on knowledge and data engineering},
  22\penalty0 (10):\penalty0 1345--1359, 2009.

\bibitem[Parisotto et~al.(2015)Parisotto, Ba, and
  Salakhutdinov]{parisotto2015actor}
E.~Parisotto, J.~L. Ba, and R.~Salakhutdinov.
\newblock Actor-mimic: Deep multitask and transfer reinforcement learning.
\newblock \emph{arXiv preprint arXiv:1511.06342}, 2015.

\bibitem[Puterman(1990)]{puterman1990markov}
M.~L. Puterman.
\newblock Markov decision processes.
\newblock \emph{Handbooks in operations research and management science},
  2:\penalty0 331--434, 1990.

\bibitem[Rusu et~al.(2016)Rusu, Rabinowitz, Desjardins, Soyer, Kirkpatrick,
  Kavukcuoglu, Pascanu, and Hadsell]{rusu2016progressive}
A.~A. Rusu, N.~C. Rabinowitz, G.~Desjardins, H.~Soyer, J.~Kirkpatrick,
  K.~Kavukcuoglu, R.~Pascanu, and R.~Hadsell.
\newblock Progressive neural networks.
\newblock \emph{arXiv preprint arXiv:1606.04671}, 2016.

\bibitem[Sabatelli et~al.(2018{\natexlab{a}})Sabatelli, Kestemont, Daelemans,
  and Geurts]{sabatelli2018deep}
M.~Sabatelli, M.~Kestemont, W.~Daelemans, and P.~Geurts.
\newblock Deep transfer learning for art classification problems.
\newblock In \emph{Proceedings of the European Conference on Computer Vision
  (ECCV) Workshops}, pages 631--646, 2018{\natexlab{a}}.

\bibitem[Sabatelli et~al.(2018{\natexlab{b}})Sabatelli, Louppe, Geurts, and
  Wiering]{sabatelli2018deepqv}
M.~Sabatelli, G.~Louppe, P.~Geurts, and M.~Wiering.
\newblock Deep quality-value (dqv) learning.
\newblock In \emph{Advances in Neural Information Processing Systems, Deep
  Reinforcement Learning Workshop}. Montreal, 2018{\natexlab{b}}.

\bibitem[Sabatelli et~al.(2020)Sabatelli, Louppe, Geurts, and
  Wiering]{sabatelli2020deep}
M.~Sabatelli, G.~Louppe, P.~Geurts, and M.~A. Wiering.
\newblock The deep quality-value family of deep reinforcement learning
  algorithms.
\newblock In \emph{2020 International Joint Conference on Neural Networks
  (IJCNN)}, pages 1--8. IEEE, 2020.

\bibitem[Sasso et~al.(2021)Sasso, Sabatelli, and Wiering]{sasso2021fractional}
R.~Sasso, M.~Sabatelli, and M.~A. Wiering.
\newblock Fractional transfer learning for deep model-based reinforcement
  learning.
\newblock \emph{arXiv preprint arXiv:2108.06526}, 2021.

\bibitem[Schulman et~al.(2015{\natexlab{a}})Schulman, Levine, Abbeel, Jordan,
  and Moritz]{schulman2015trust}
J.~Schulman, S.~Levine, P.~Abbeel, M.~Jordan, and P.~Moritz.
\newblock Trust region policy optimization.
\newblock In \emph{International conference on machine learning}, pages
  1889--1897. PMLR, 2015{\natexlab{a}}.

\bibitem[Schulman et~al.(2015{\natexlab{b}})Schulman, Moritz, Levine, Jordan,
  and Abbeel]{schulman2015high}
J.~Schulman, P.~Moritz, S.~Levine, M.~Jordan, and P.~Abbeel.
\newblock High-dimensional continuous control using generalized advantage
  estimation.
\newblock \emph{arXiv preprint arXiv:1506.02438}, 2015{\natexlab{b}}.

\bibitem[Schulman et~al.(2017)Schulman, Wolski, Dhariwal, Radford, and
  Klimov]{schulman2017proximal}
J.~Schulman, F.~Wolski, P.~Dhariwal, A.~Radford, and O.~Klimov.
\newblock Proximal policy optimization algorithms.
\newblock \emph{arXiv preprint arXiv:1707.06347}, 2017.

\bibitem[Sharif~Razavian et~al.(2014)Sharif~Razavian, Azizpour, Sullivan, and
  Carlsson]{sharif2014cnn}
A.~Sharif~Razavian, H.~Azizpour, J.~Sullivan, and S.~Carlsson.
\newblock Cnn features off-the-shelf: an astounding baseline for recognition.
\newblock In \emph{Proceedings of the IEEE conference on computer vision and
  pattern recognition workshops}, pages 806--813, 2014.

\bibitem[Taylor and Stone(2009)]{taylor2009transfer}
M.~E. Taylor and P.~Stone.
\newblock Transfer learning for reinforcement learning domains: A survey.
\newblock \emph{Journal of Machine Learning Research}, 10\penalty0 (7), 2009.

\bibitem[Tirinzoni et~al.(2018)Tirinzoni, Sanchez, and
  Restelli]{tirinzoni2018transfer}
A.~Tirinzoni, R.~R. Sanchez, and M.~Restelli.
\newblock Transfer of value functions via variational methods.
\newblock In \emph{Advances in Neural Information Processing Systems}, pages
  6179--6189, 2018.

\bibitem[Tyo and Lipton(2020)]{tyo2020transferable}
J.~Tyo and Z.~Lipton.
\newblock How transferable are the representations learned by deep q agents?
\newblock \emph{arXiv preprint arXiv:2002.10021}, 2020.

\bibitem[van~de Wolfshaar(2017)]{vanjos2017deep}
J.~van~de Wolfshaar.
\newblock \emph{Deep Reinforcement Learnig of Video Games}.
\newblock PhD thesis, Faculty of Science and Engineering, 2017.

\bibitem[van~de Wolfshaar et~al.(2018)van~de Wolfshaar, Wiering, and
  Schomaker]{vanjos2018deep}
J.~van~de Wolfshaar, M.~A. Wiering, and L.~Schomaker.
\newblock Deep learning policy quantization.
\newblock 2018.

\bibitem[Van~Hasselt et~al.(2016)Van~Hasselt, Guez, and Silver]{van2016deep}
H.~Van~Hasselt, A.~Guez, and D.~Silver.
\newblock Deep reinforcement learning with double {Q}-learning.
\newblock In \emph{Thirtieth AAAI Conference on Artificial Intelligence}, 2016.

\bibitem[Vandaele et~al.(2021)Vandaele, Dance, and Ojha]{vandaele2021deep}
R.~Vandaele, S.~L. Dance, and V.~Ojha.
\newblock Deep learning for automated river-level monitoring through
  river-camera images: an approach based on water segmentation and transfer
  learning.
\newblock \emph{Hydrology and Earth System Sciences}, 25\penalty0 (8):\penalty0
  4435--4453, 2021.

\bibitem[Wang et~al.(2016{\natexlab{a}})Wang, Bapst, Heess, Mnih, Munos,
  Kavukcuoglu, and de~Freitas]{wang2016sample}
Z.~Wang, V.~Bapst, N.~Heess, V.~Mnih, R.~Munos, K.~Kavukcuoglu, and
  N.~de~Freitas.
\newblock Sample efficient actor-critic with experience replay.
\newblock \emph{arXiv preprint arXiv:1611.01224}, 2016{\natexlab{a}}.

\bibitem[Wang et~al.(2016{\natexlab{b}})Wang, Schaul, Hessel, Van~Hasselt,
  Lanctot, and Freitas]{wang2016dueling}
Z.~Wang, T.~Schaul, M.~Hessel, H.~Van~Hasselt, M.~Lanctot, and N.~Freitas.
\newblock Dueling network architectures for deep reinforcement learning.
\newblock In \emph{International Conference on Machine Learning}, pages
  1995--2003, 2016{\natexlab{b}}.

\bibitem[Zhao et~al.(2016)Zhao, Wang, Shao, and Zhu]{zhao2016deep}
D.~Zhao, H.~Wang, K.~Shao, and Y.~Zhu.
\newblock Deep reinforcement learning with experience replay based on sarsa.
\newblock In \emph{2016 IEEE Symposium Series on Computational Intelligence
  (SSCI)}, pages 1--6. IEEE, 2016.

\bibitem[Zhu et~al.(2017)Zhu, Mottaghi, Kolve, Lim, Gupta, Fei-Fei, and
  Farhadi]{zhu2017target}
Y.~Zhu, R.~Mottaghi, E.~Kolve, J.~J. Lim, A.~Gupta, L.~Fei-Fei, and A.~Farhadi.
\newblock Target-driven visual navigation in indoor scenes using deep
  reinforcement learning.
\newblock In \emph{2017 IEEE international conference on robotics and
  automation (ICRA)}, pages 3357--3364. IEEE, 2017.

\bibitem[Zhuang et~al.(2020)Zhuang, Qi, Duan, Xi, Zhu, Zhu, Xiong, and
  He]{zhuang2020comprehensive}
F.~Zhuang, Z.~Qi, K.~Duan, D.~Xi, Y.~Zhu, H.~Zhu, H.~Xiong, and Q.~He.
\newblock A comprehensive survey on transfer learning.
\newblock \emph{Proceedings of the IEEE}, 109\penalty0 (1):\penalty0 43--76,
  2020.

\end{thebibliography}

\appendix
\section{Appendix}

\label{sec:experimental_setup_control}

We hereafter provide additional information about how the model-free DRL algorithms considered throughout this paper were trained. In Table \ref{tab:table_ddqn} and \ref{tab:table_dqv}, we provide the hyper-parameters that were used for the large scale empirical study presented in Sec. \ref{sec:empirical_study}, while we then describe the DQN architecture used for the control experiments presented in Sec. \ref{sec:control_experiments}.

\begin{table}[ht!]
\centering
\caption{Hyper-parameters used when training a DDQN agent from scratch. We follow the experimental setup introduced in the original paper \cite{van2016deep}.}
\begin{tabular}{c | c |  }
Hyper-parameter \\
\hline \hline
Atari Arcade Learning Version  & \texttt{Deterministic-v4} \\
Frame-Skipping & True \\ 
Reward Clipping & $[-1,1]$ \\ 
Epsilon Greedy $\epsilon$ & 0.1 \\
Discount Factor $\gamma$ & 0.99 \\ 
Pre-processing scheme & $84\times84\times4$ \\ 
$Q$-optimizer & RMSprop \\ 
$Q$ Learning rate & 0.00025 \\ 
Optimizer $\rho$ & 0.95 \\ 
Optimizer $\epsilon$ & 0.01 \\ 
Memory size $S$ & 1M trajectories 
\end{tabular}
\label{tab:table_ddqn}
\end{table}

\begin{table}[ht!]
\centering
\caption{Hyper-parameters used when training a DQV agent from scratch. We can see that they mostly correspond to the ones presented in Table \ref{tab:table_ddqn} with the main difference being the epsilon greedy parameter $\epsilon$ that is set to 0.5 instead of 1.0, and the additional information given for the $V$ network which is trained for learning an approximation of the state-value function.}
\begin{tabular}{c | c |  }
Hyper-parameter \\
\hline \hline
Atari Arcade Learning Version  & \texttt{Deterministic-v4} \\
Frame-Skipping & True \\ 
Reward Clipping & $[-1,1]$ \\ 
Epsilon Greedy $\epsilon$ & 0.5 \\
Discount Factor $\gamma$ & 0.99 \\ 
Pre-processing scheme & $84\times84\times4$ \\ 
$Q$-optimizer & RMSprop \\ 
$Q$ Learning rate & 0.00025 \\ 
$V$-optimizer & SGD \\ 
$V$ Learning rate & 0.001 \\
Optimizer $\rho$ & 0.95 \\ 
Optimizer $\epsilon$ & 0.01 \\ 
Memory size $S$ & 1M trajectories 
\end{tabular}
\label{tab:table_dqv}
\end{table}

\paragraph{DQN:} The agent comes in the form a two-hidden layer convolutional neural network which is followed by a fully connected layer of $256$ hidden units preceding the final linear layer responsible for estimating the different $Q(s,a)$ values. The first convolutional layer has $32$ channels whereas the second one has $64$ channels. All layers of the network are activated by a ReLU non-linearity. We use an experience replay memory buffer that is set to contain $400000$ trajectories, and start training the model as soon as $5000$ trajectories have been collected. For exploration purposes, we adopt the popular epsilon-greedy strategy with an initial $\epsilon$ value of $1$ which gets linearly annealed over time to $0.1$. Learning a near optimal policy on the aforementioned \texttt{Catch} environments with this type of DQN agent can require between the $3$ and the $5$ hours of training time.

\end{document}